\documentclass[acmtog]{acmart}
\usepackage{bm}

\AtBeginDocument{%
  }

\setcopyright{rightsretained}
\copyrightyear{2024}
\acmYear{2024}
\acmDOI{XXXXXXX.XXXXXXX}

\usepackage{bbding}
\usepackage{makecell}
\usepackage{multirow}
\usepackage{array}
\usepackage{subfigure}
\usepackage{booktabs}
\definecolor{mycolor}{HTML}{D8ECD1}

\definecolor{baselinecolor}{gray}{.9}

\newlength\savewidth

\definecolor{dt}{gray}{0.6}  %
\newcolumntype{*}{>{\global\let\currentrowstyle\relax}}
\newcolumntype{^}{>{\currentrowstyle}}
\newcommand{\rowstyle}[1]{\gdef\currentrowstyle{#1}#1\ignorespaces}

\makeatletter\renewcommand\paragraph{\@startsection{paragraph}{4}{\z@}
  {.5em \@plus1ex \@minus.2ex}{-.5em}{\normalfont\normalsize\bfseries}}\makeatother

\newcolumntype{x}[1]{>{\centering\arraybackslash}p{#1pt}}
\newcolumntype{y}[1]{>{\raggedright\arraybackslash}p{#1pt}}
\newcolumntype{z}[1]{>{\raggedleft\arraybackslash}p{#1pt}}

\begin{document}

\title{High-Fidelity Facial Albedo Estimation via Texture Quantization}

\author{Zimin Ran}
\affiliation{%
 \institution{University of Technology Sydney}
 \city{Sydney}
 \country{Australia}}
\email{zimin.ran@student.uts.edu.au}

\author{Xingyu Ren}
\affiliation{%
 \institution{Shanghai Jiao Tong University}
 \city{Shanghai}
 \country{China}}
\email{rxy_sjtu@sjtu.edu.cn}

\author{Xiang An}
\affiliation{%
 \institution{DeepGlint}
 \city{Beijing}
 \country{China}}

\author{Kaicheng Yang}
\affiliation{%
 \institution{DeepGlint}
 \city{Beijing}
 \country{China}}

\author{Xiangzi Dai}
\affiliation{%
 \institution{DeepGlint}
 \city{Beijing}
 \country{China}}

\author{Ziyong Feng}
\affiliation{%
 \institution{DeepGlint}
 \city{Beijing}
 \country{China}}

\author{Jia Guo}
\affiliation{%
 \institution{Insightface}
 \city{Shanghai}
 \country{China}}

\author{Linchao Zhu}
\affiliation{%
 \institution{Zhejiang University}
 \city{Hangzhou}
 \country{China}}

\author{Jiankang Deng}
\affiliation{%
 \institution{Imperial College London}
 \city{London}
 \country{United Kingdom}}
\email{jiankangdeng@gmail.com}

\renewcommand{\shortauthors}{Ran, Z. et al.}

\begin{abstract}
Recent 3D face reconstruction methods have made significant progress in shape estimation, but high-fidelity facial albedo reconstruction remains challenging. Existing methods depend on expensive light-stage captured data to learn facial albedo maps. However, a lack of diversity in subjects limits their ability to recover high-fidelity results. In this paper, we present a novel facial albedo reconstruction model, HiFiAlbedo, which recovers the albedo map directly from a single image without the need for captured albedo data. Our key insight is that the albedo map is the illumination invariant texture map, which enables us to use inexpensive texture data to derive an albedo estimation by eliminating illumination. To achieve this, we first collect large-scale ultra-high-resolution facial images and train a high-fidelity facial texture codebook. By using the FFHQ dataset and limited UV textures, we then fine-tune the encoder for texture reconstruction from the input image with adversarial supervision in both image and UV space. Finally, we train a cross-attention module and utilize group identity loss to learn the adaptation from facial texture to the albedo domain. Extensive experimentation has demonstrated that our method exhibits excellent generalizability and is capable of achieving high-fidelity results for in-the-wild facial albedo recovery. Our code, pre-trained weights, and training data will be made publicly available at \textbf{\url{https://hifialbedo.github.io/}}.
\end{abstract}



\begin{teaserfigure}
  \includegraphics[width=\textwidth]{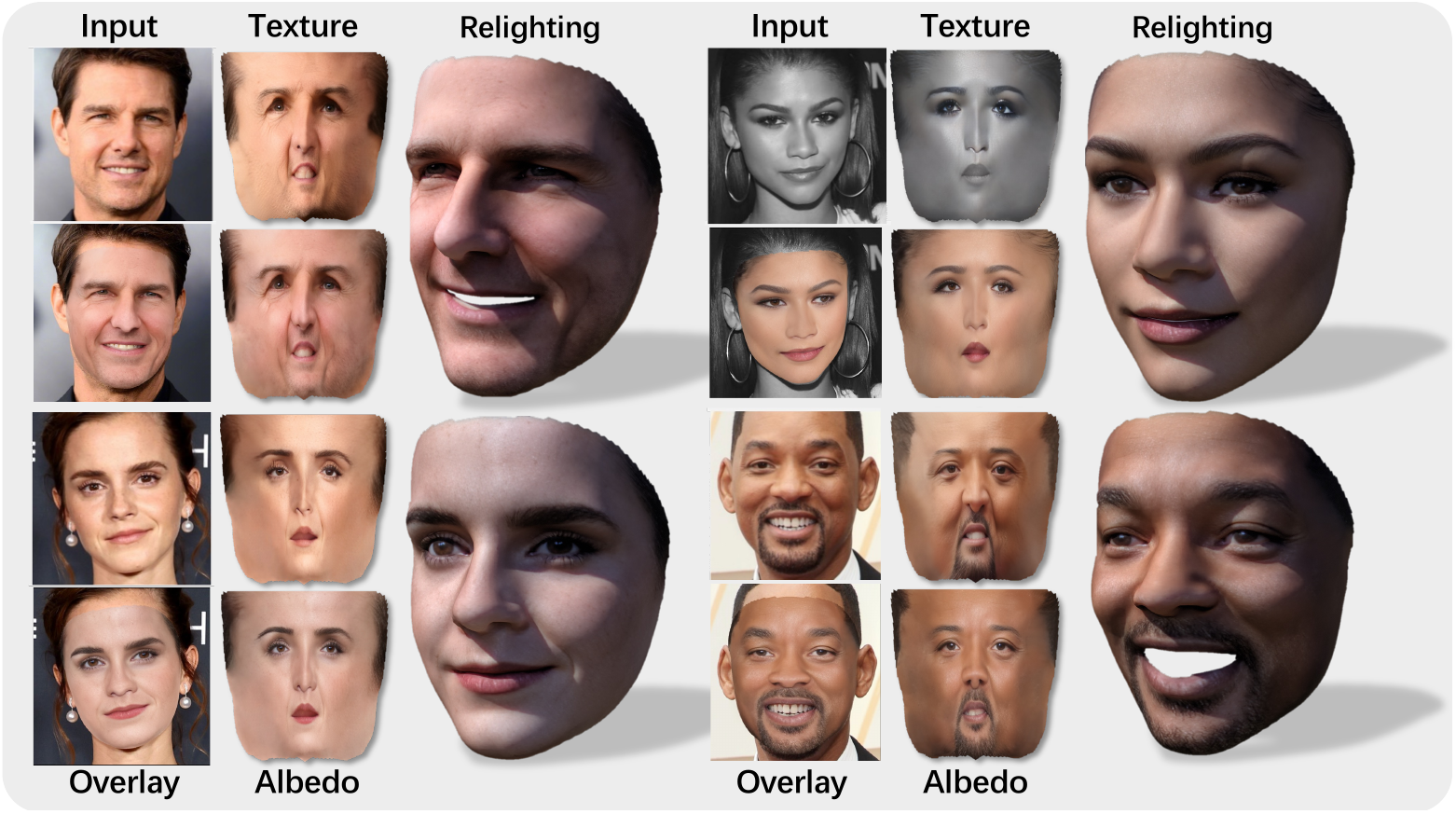}
  \caption{We introduce HiFiAlbedo, a high-fidelity facial albedo recovery method. HiFiAlbedo learns a high-quality texture codebook from large-scale RGB faces, generates textures from the images, and then enables domain adaptation from texture to the albedo domain. Our method does not rely on captured data and generates high-fidelity albedo maps that can be used for realistic rendering.}
  \label{fig:teaser}
\end{teaserfigure}


\maketitle

\section{Introduction}
3D face reconstruction is a prominent and challenging area of research in the domains of computer vision and graphics~\cite{tewari2022advances,Khakhulin2022realistic}. It has garnered significant attention from researchers due to its versatile applications across diverse multimedia domains, including virtual reality (VR) and augmented reality (AR). 3D face reconstruction provides a foundation for the creation of avatars~\cite{zhang2023dreamface} in the metaverse or video games, facilitates virtual social interactions, and supports advancements in plastic surgery, among other applications. To achieve more realistic results, high-fidelity albedo recovery has garnered significant attention.

To produce high-fidelity facial albedo maps, the industrial pipeline typically employs professional equipment, e.g., Light Stage~\cite{debevec2012light}, to capture high-quality multi-view polarized images. Then photo-metric stereo~\cite{ghosh2011multiview} in combination with monochrome color reconstruction using polarization promotion~\cite{LeGendre2018facial} enables the capture of albedo maps with a high degree of accuracy at the pore level. As this process necessitates a subsequent manual post-processing step by artists, the collection of albedo data is a costly endeavor, with fewer than 200 subject albedos currently available for purchase.

Given limited albedo data, existing methods employ two main approaches to generate high-quality albedo. The first approach~\cite{deng2019accurate,smith2020morphable,feng2022towards,ren2023facial} involves building a data-based prior with real albedo data, such as a statistical model (e.g., PCA) or a generative model (e.g., StyleGAN~\cite{karras2019style}). The albedo model can be conditioned by latent code to obtain new albedo maps. However, these methods only recover approximate skin colors, making it difficult to achieve high-fidelity albedo recovery.
The second approach first reconstructs the face texture and then trains an image-to-image transformation network to transfer the texture to the albedo domain. However, these approaches~\cite{Lattas20,lattas2021avatarme++} struggle to generalize well to diverse real-world identities. Recent works such as FitMe~\cite{lattas2023fitme} and Relightify~\cite{papantoniou2023relightify} have shown impressive performance in facial albedo reconstruction by learning large-scale facial reflection prior by using StyleGAN~\cite{karras2019style} and the latent diffusion model~\cite{rombach2022high}. Despite this, these works~\cite{Lattas20,lattas2021avatarme++,lattas2023fitme,papantoniou2023relightify} have not made any facial albedo data publicly available.

\begin{figure}[t]
    \centering
    \includegraphics[width=\linewidth]{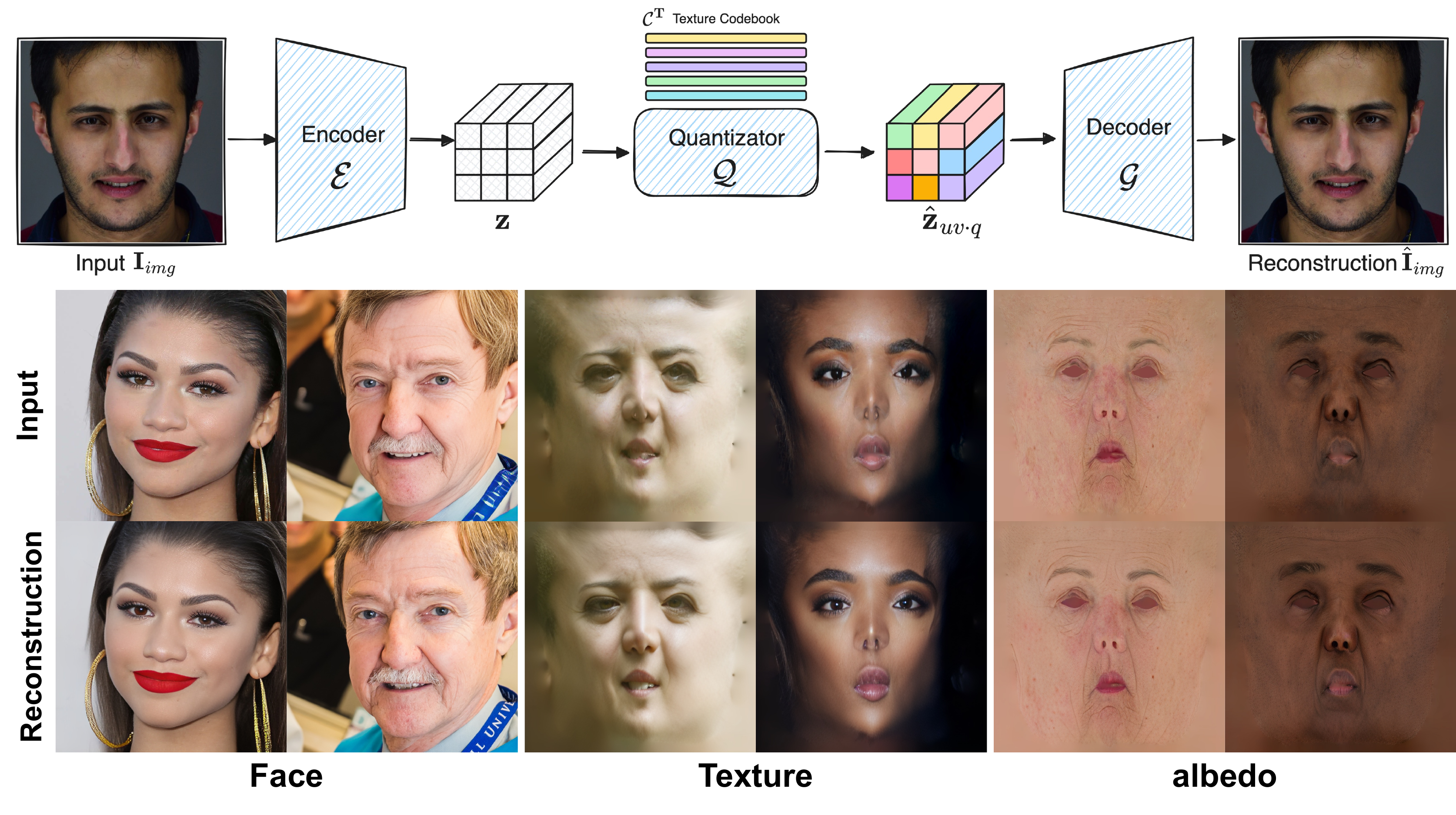}
    \caption{Visualization of the reconstruction results based on the codebook. Given facial images, textures, or albedos as inputs, our pre-trained VQGAN consistently achieves high-fidelity reconstruction results.}
    \label{fig:3_1}
\end{figure}

Since the albedo map represents the intrinsic face color independent of the extrinsic lighting conditions, could we start from a high-quality texture and achieve de-lighting effects by domain adaptation? In this paper, we provide an affirmative answer by introducing a novel HiFiAlbedo model.
Specifically, we first train a high-fidelity face texture prior. Due to the scarcity of high-fidelity UV texture data, texture model training is not easy. Since we experimentally found that the image structure and codebook are deconstructed in the vector quantization-based generation model (shown in Fig.~\ref{fig:3_1}), we train a face texture codebook directly in image space to reconstruct UV texture.
To obtain UV textures from input face images, we fine-tune the original encoder using a combination of FFHQ data and limited texture data. However, due to the large distribution gap between these latent codes, there are noticeable artifacts during the reconstruction process. Therefore, we design a dual discrimination module that jointly regulates the latent codebook distribution and the RGB image. Once fine-tuned, this model can produce a high-quality UV texture from any input face image.

Based on the texture reconstruction, we further achieve albedo estimation from the predicted UV textures.
We consider albedo estimation as a Monte Carlo sampling process. In that case, the average albedo estimation aligns with the ground truth albedo, provided that the samples are images of the same person in various scenes under different lighting conditions.
Specifically, we initialize a learnable query latent and introduce a cross-attention module optimized by the group identity loss. This training process teaches cross-attention from various facial textures to the same albedo latent variable. After training, our model can accurately infer the corresponding albedo map from a single face input, as illustrated in Fig.~\ref{fig:teaser} and Fig.~\ref{fig:3_3}. The results demonstrate that our method consistently produces high-fidelity albedos and effectively generalizes to diverse inputs.

In summary, the main contributions of this work are as follows:
\begin{itemize}
    \item We propose a novel high-fidelity albedo recovery model, HifiAlbedo, that utilizes multiple facial images of the same individual, eliminating the need for data capture.
    \item We propose a dual discriminator, which achieves high-quality texture prediction from a single image by joint discrimination in latent space and image space.
    \item We design a group identity loss that enforces the generation of realistic and identity-consistent facial albedos from a set of images under different illuminations.
    \item The proposed HifiAlbedo method significantly improves the fidelity of facial albedo estimation and achieves competitive results in the FAIR benchmark.
\end{itemize}

\begin{figure*}[t]
    \centering
    \includegraphics[width=\linewidth]{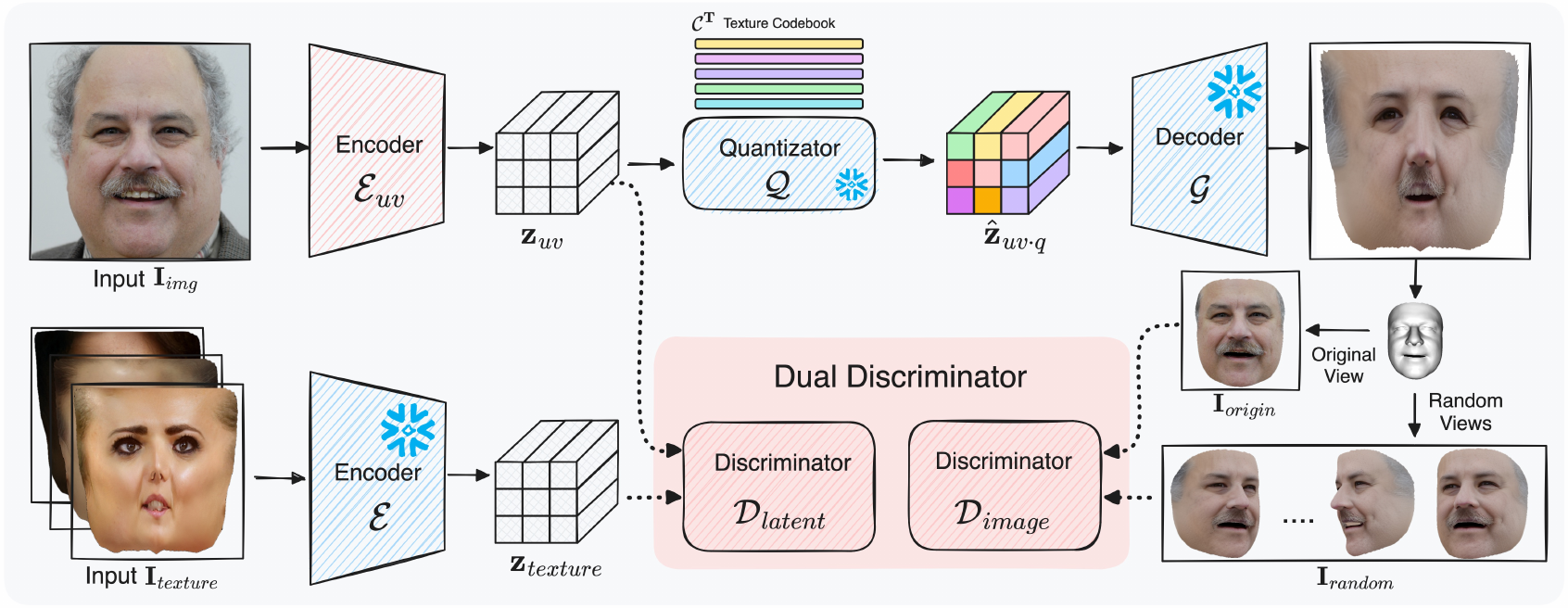}
    \caption{Overview of our UV texture reconstruction pipeline. After training a VQ-based auto-encoder (blue box), we fine-tune the encoder and propose a dual discriminator (pink box). UV texture reconstruction from a single image is achieved by adversarial supervision in both latent and image space.}
    \label{fig:texture}
\end{figure*}

\section{Related Work}
\noindent{\bf Facial Albedo Reconstruction.}
Current approaches to monocular face reconstruction predominantly use statistical face models such as the 3D Morphable Model (3DMM). This model includes a geometric space for shape reconstruction and an appearance space for albedo reconstruction~\cite{Egger2020}. One of the most widely used models is the Basel Face Model (BFM)~\cite{Paysan09}, which was derived from a dataset of approximately 200 European subjects. The limited diversity in this dataset has resulted in a biased appearance space. 
To address this critical issue, Smith et al.~\cite{smith2020morphable} introduced AlbedoMM, which innovatively constructs an albedo model by leveraging diverse light-stage data and simultaneously modeling specular and diffuse albedo to enrich diversity within the appearance space.

In parallel to PCA-based models, GAN-based approaches have also gained prominence. Deng et al.~\cite{deng2018uv} pioneered UV texture completion for in-the-wild face images.
Gecer et al.~\cite{Gecer19:ganfit,gecer2021fast} employed a texture GAN trained on a large dataset of 10K textures, significantly enhancing realism. However, their approach resulted in textures with baked illumination.
Lattas et al.~\cite{Lattas20,lattas2021avatarme++} employed an image-to-image translation network, trained on light-stage data, to generate both diffuse and specular albedo from high-quality textures. However, the training data limits its ability to generalize across diverse racial groups. 
Our research overcomes the reliance on data capture, culminating in the creation of a high-resolution albedo generator by using large-scale high-quality web face images.

\noindent{\bf Light and Albedo Ambiguity.}
Extracting accurate illumination and albedo from image appearances is a complex, ill-defined challenge~\cite{Ramamoorthi01}. While applying appearance priors has helped to limit variations in albedo, it hasn't fully resolved the issue of ambiguity. The common strategy involves employing more robust priors to constrain both aspects. Hu et al.~\cite{Hu13} proposed a method to normalize albedo symmetry. Aldrian et al.~\cite{Aldrian12} introduced a regularization technique for lighting by applying a ``gray world'' assumption, which ensures the light appears monochromatic. Following this, related studies~\cite{deng2019accurate,Feng2021}, adopted a similar methodology for regularization to facilitate an estimated decomposition. Egger et al.~\cite{Egger18} considered the specific distribution of illumination, and developed a statistical prior directly for the Spherical Harmonics (SH) coefficients. TRUST~\cite{feng2022towards} expanded the capabilities of light estimation by dividing light into two categories: face light and ambient light, and exploited the consistency of ambient light to improve the accuracy of light estimation. FFHQ-UV~\cite{bai2023ffhq} leveraged StyleGAN to normalize lighting conditions during the pre-processing of the dataset. ID2Albedo ~\cite{ren2023improving} utilized a variety of facial attribute priors to refine the albedo generation. Recently ID2Reflectance~\cite{ren2024monocular} apply face swapping technology to reflectance domain and achieve high quality facial albedo generation. Departing from these regularization methods, our approach presents a novel strategy by introducing a shared albedo latent space combined with a group identity loss, allowing unsupervised domain adaptation from texture to albedo.

\section{Methodology}
This work aims to reconstruct high-fidelity facial texture and albedo from a single in-the-wild image.
To this end, we first learn a high-fidelity facial texture codebook in the image space~(Sec.~\ref{Sec:Codebooklearning}). 
Based on the pre-trained texture codebook, we design the dual discrimination modules (Fig.~\ref{fig:texture}) to support facial image unwrapping and reconstruct photo-realistic UV texture (Sec.~\ref{Sec:Texture}).
Finally, we propose a latent attention module (Fig.~\ref{fig:3_3}) with group ID loss to achieve high-fidelity facial albedo map reconstruction (Sec.~\ref{Sec:Albedo}). 

\noindent{\bf Key Idea.} The biggest challenge in building expressive facial albedo models is the lack of large-scale, high-quality albedo maps that are collected from diverse identities. 
Existing albedo reconstruction methods~\cite{deng2019accurate,smith2020morphable,feng2022towards,ren2023improving} learn from scratch using limited captured data in the facial UV space, but these solutions lead to a lack of generalization in the albedo domain. 
Therefore, we divide albedo reconstruction into three steps: (1) constructing a high-quality face texture prior from large-scale inexpensive face RGB data, (2) fine-tuning the encoder to achieve UV texture reconstruction for the input face image, and (3) achieving domain adaptation from texture to albedo using group identity constraints.

\subsection{Facial Texture Codebook Learning}
\label{Sec:Codebooklearning}
Since recent quantized auto-encoders~\cite{van2017neural} enable the deconstruction of image structures and codebooks, we first learn a high-fidelity facial texture codebook~\cite{esser2021taming} in the image space. 
To learn photo-realistic texture with pore-level skin details, we filter a large amount of high-resolution face data and collect half a million unaligned face images from the WebFace260M celebrity dataset\cite{zhu2021webface260m}.
All these data are cropped into $1024\times1024$ resolution with the template used by Arcface~\cite{deng2019arcface,deng2020retinaface} to ensure a higher percentage of facial regions in the image.

Given the input face image $\mathbf{I} \in \mathbb{R}^{H \times W \times 3}$, it is embedded by the encoder $\mathcal{E}$ into a feature embedding $\mathbf{z} \in \mathbb{R}^{h \times w \times d}$. 
To effectively learn a latent discrete spatial codebook, we use nearest-neighbor search to get the nearest item and quantized feature from the codebook $\mathcal{C}=\left\{c_n \in \mathbb{R}^{d}\right\}_{n=0}^N$:
\begin{equation}
\mathbf{z}^{(i,j)}_q = \mathcal{Q}({\mathbf{z}}^{(i,j)}) := \arg\min_{{c_n} \in \mathcal{C}} \Vert {{z}^{(i,j)}} - {c_n} \Vert,
\label{eq:vq}
\end{equation}
where $\mathbf{z}_{q} \in \mathbb{R}^{{h} \times {w} \times {d}}$ is the quantized feature, $\mathcal{Q}(\cdot)$ is a quantization operation, and $N=1,024$ is the number of code items in the codebook. Taking the quantized representation $z_{q}$ as input, the decoder $\mathcal{G}$ can reconstruct the high-quality face image $\hat{\mathbf{I}}=\mathcal{G}(\mathcal{Q}(\mathcal{E}(\mathbf{I})))$).
The reconstruction loss $\mathcal{L}_{rec}$ is defined as $\mathcal{L}_{rec} = \left\|\mathbf{I}-\hat{\mathbf{I}}\right\|^2.$
Following the previous work~\cite{van2017neural,esser2021taming}, we also adopt commitment loss to reduce the distance between the quantized feature $\mathbf{z}_q$ and the feature embedding $\mathbf{z}$:
\begin{equation}
\mathcal{L}_{com} = \left \| \mathrm{sg}\left [ \mathcal{E}(\mathbf{I}) \right ] - {\mathbf{z}_q} \right \|_{2}^{2} + \beta \left \| \mathrm{sg}\left [ {\mathbf{z}_q} \right ] - \mathcal{E}(\mathbf{I}) \right \|_{2}^{2},
\label{eq:code}
\end{equation}
where $\mathrm{sg}(\cdot)$ denotes the stop-gradient operator and $\beta$ is a weight factor. To further reduce the difference between reconstructed textures and input textures, we introduce an adversarial training procedure with a patch-based discriminator $\mathcal{D}$, and the adversarial loss $\mathcal{L}_{adv1}$ is defined as:
\begin{equation}
\mathcal{L}_{adv1} = log \mathcal{D}(\mathbf{I}) + log(1-\mathcal{D}(\hat{\mathbf{I}})).
\label{eq:adv}
\end{equation}
Finally, the overall optimization objective $\mathcal{L}_{pretrain}$ in the codebook pretraining process is:
\begin{equation}
\mathcal{L}_{pretrain} = \mathcal{L}_{rec} + \mathcal{L}_{com} + \lambda_{0} \cdot \mathcal{L}_{adv1},
\end{equation}
where the loss weights $\lambda_{0}$ is set as 0.8.

\subsection{UV Texture Reconstruction}
\label{Sec:Texture}
\begin{figure*}[t] 
    \centering 
    \includegraphics[width=\linewidth]{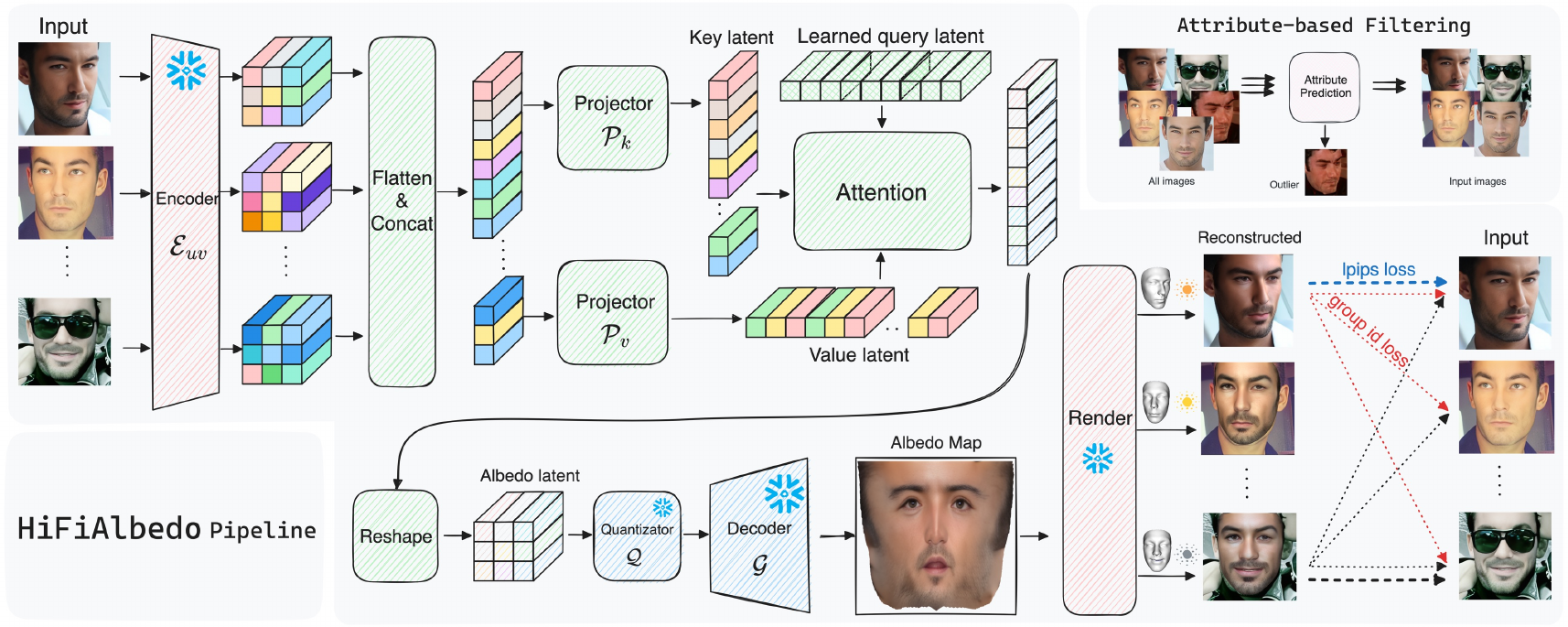} 
    \caption{Overview of the proposed method HiFiAlbedo. Our core insight is that an individual shares the same albedo in different scenes. Therefore, we propose a group identity loss for unsupervised domain adaptation from texture to albedo. Specifically, we first sample the faces of the same person with similar attributes as input and then extract features by using the encoder trained in Sec.~\ref{Sec:Texture}. These features are projected separately, and then cross-attention is computed using the learnable query latent to obtain the shared albedo. Finally, the albedo is overlaid back onto the original image, and then the group identity loss is computed to obtain a high-fidelity face albedo.} 
    \label{fig:3_3} 
\end{figure*}
After training in Sec.~\ref{Sec:Codebooklearning}, we have built a high-quality texture codebook. However, this codebook only supports homography reconstruction of the input face images or UVs. To enable texture unwrapping for any input face image, we fine-tune the encoder $\mathcal{E}$ from the first step. Specifically, we propose a dual-discriminator combined with our auto-encoder framework, as shown in Fig.~\ref{fig:texture}. This dual discriminator design constrains the distribution in both code and image space. The code discriminator acts before extracting the discrete code representation to ensure topological consistency. However, using it alone is not sufficient. We notice many imperfections in unseen regions, as shown in Fig.~\ref{fig:dualdis}. To this end, we use an image discriminator to ensure the correctness of the novel view regions. We utilize the off-the-shelf face reconstruction model to generate a 3D facial mesh from the original image, map the generated texture map onto the mesh, render it from the original and random viewpoints, and feed it into the image discriminator. Our dual discriminator design enforces the encoder to produce consistent and realistic UV texture mappings.

To fine-tune this encoder, we employ the original FFHQ dataset and limited UV texture data~\cite{bai2023ffhq}.
For a given image input $\mathbf{I}_{img}$, we first obtain the feature embedding $\mathbf{z}_{uv}$, by fine-tuning the encoder $\mathcal{E}_{uv}$. Similarly, we get the embedding $\mathbf{z}_{texture}$ from the UV texture input $\mathbf{I}_{texture}$. After quantization, we obtain $\mathbf{\hat z}_{uv \cdot q}$ and the final texture output $\mathbf{T}=\mathcal{G}(\mathcal{Q}(\mathcal{E}_{uv}(\mathbf{z}_{uv \cdot q})))$. We then map this texture onto the predicted mesh and render $\mathbf{I}_{origin}$ and $\mathbf{I}_{random}$ from the original and random viewpoints, respectively.

During training, we fix the pre-trained codebook and decoder and optimize the encoder and dual discriminator.
We employ the identity loss function, which is the cosine distance between the input face $\mathbf{I}_{img}$ and the random view $\mathbf{I}_{random}$:
\begin{equation}
\mathcal{L}_{id} =  1 - \frac{\mathcal{A}(\mathbf{I}_{img})\mathcal{A}(\mathbf{I}_{random})}{\|\mathcal{A}(\mathbf{I}_{img})\|_2\cdot \|\mathcal{A}(\mathbf{I}_{random})\|_2},
\label{eq:identityloss}
\end{equation}
where $\mathcal{A}$ is the pre-trained ArcFace model~\cite{deng2019arcface}. Besides, the adversarial loss of our dual discriminator can be calculated as follows:
\begin{equation}
\begin{aligned}
\mathcal{L}_{adv2} &= log \mathcal{D}_{latent}(\mathbf{z}_{uv}) + log(1-\mathcal{D}_{latent}(\mathbf{z}_{texture})) \\ &+ log \mathcal{D}_{image}(\mathbf{I}_{origin}) + log(1-\mathcal{D}_{image}(\mathbf{I}_{random})),
\end{aligned}
\label{eq:GAN}
\end{equation}
where $\mathcal{D}_{latent}$ and $\mathcal{D}_{image}$ are our dual discriminators. The complete training goal is as follows:
\begin{equation}
\mathcal{L}_{texture} = \mathcal{L}_{id} + \lambda_{1}\mathcal{L}_{rec} + \lambda_{2}\mathcal{L}_{lpips} + \lambda_{3}\mathcal{L}_{adv2},
\label{eq:all}
\end{equation}
where $\mathcal{L}_{rec}$ denotes the L1 loss between the input $\mathbf{I}_{img}$ and the warped $\mathbf{I}_{origin}$.
$\mathcal{L}_{lpips}$ denotes the LPIPS loss~\cite{zhang2018perceptual}.
The loss weights $\lambda_{1}$, $\lambda_{2}$, $\lambda_{3}$ are set as $10$, $10$ and $0.1$, respectively.

\subsection{High-fidelity Albedo Disambiguation}
\label{Sec:Albedo}
Assuming that the face is a Lambertian surface, the rendered face image can be computed by
$\mathcal{R}= \mathcal{A} \odot \mathcal{S}$, where $\mathcal{R}$ stands for the final rendered image, $\odot$ denotes the Hadamard product, and $\mathcal{A}$ and $\mathcal{S}$ represent the wrapped face albedo and the shading image, respectively. When there is a parallel estimation of both albedo and illumination, the ambiguity between albedo and illumination occurs. Recent work has been conducted by ensuring different individuals in the same scene have consistent lighting estimation~\cite{feng2022towards} or by using face attributes for direct inference~\cite{ren2023improving}.
However, they all require expensive hand-crafted albedo maps to generate training data.

Assuming that the face imaging process under any lighting condition is a stochastic process that follows an independent and identical distribution, we can estimate the facial albedo using the Monte Carlo sampling method. In other words, if we have a large number of images of the same person under different lighting conditions, our albedo estimation will gradually approach the ground truth.
To this end, we propose the novel HiFiAlbedo pipeline, which adaptively extracts a high-fidelity albedo map from multiple photos of a subject, as shown in Fig.~\ref{fig:3_3}.

Given multiple images of the same person, we first apply attribute-based filtering. Considering that photos of the same subject vary widely, we extract attributes such as the age of the person in the image to ensure accurate and consistent albedo. Our sampling constraint will prevent the outlier data and help maintain consistent albedos in each image batch during training, which will aid in the correct recovery of individual albedos, as shown in the upper right corner of Fig.~\ref{fig:3_3}.
After filtering, we feed the input images $\mathbf{I}_{1}$, $\mathbf{I}_{2}$, $\mathbf{I}_{3}$, $\dots$, $\mathbf{I}_{n}$ to a pre-trained encoder $\mathcal{E}_{uv}$ that extracts these continuous latents $\mathbf{z}_{uv \cdot 1}$, $\mathbf{z}_{uv \cdot 2}$, $\mathbf{z}_{uv \cdot 3}$, $\dots$, $\mathbf{z}_{uv \cdot n}$. To infer the albedo latent, we introduce a cross-attention module to learn the correspondence between the texture space and the albedo space.

\begin{figure*}[t!]
  \centering
  \includegraphics[width=\textwidth]{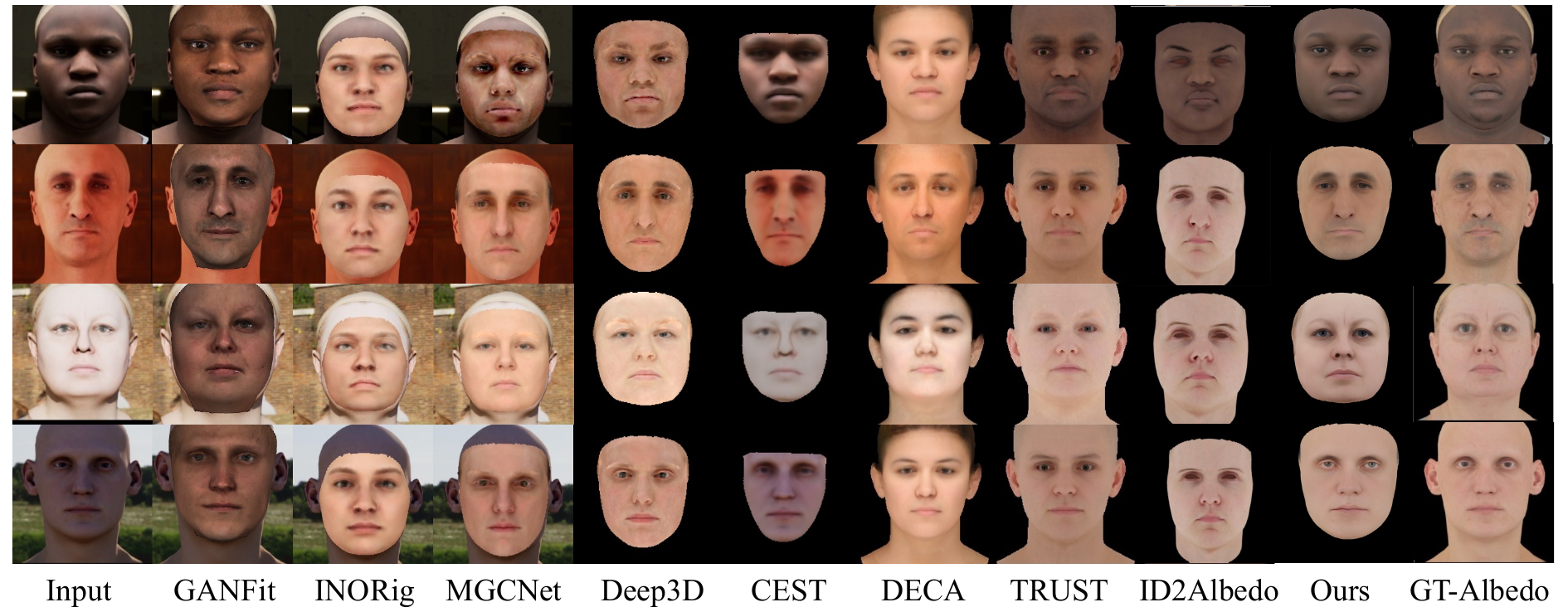}
  \caption{Comparison on the FAIR benchmark~\cite{feng2022towards}. From left to right: input image, GANFIT~\cite{Gecer19:ganfit}, INORig~\cite{Bai21}, MGCNet~\cite{shang2020self}, Deep3D~\cite{deng2019accurate}, CEST~\cite{Wen21}, DECA~\cite{Feng2021}, TRUST ~\cite{feng2022towards}, ID2Albedo~\cite{ren2023improving}, ours and ground-truth albedo rendering. }
\label{fig:fairness}
\end{figure*}

\begin{table*}[t]
\centering
\resizebox{0.95\linewidth}{!}{%
\begin{tabular}{ *l   | ^c | ^c | ^c | ^c | ^c^c^c^c^c^c }
\toprule
\multirow{2}{*}{Method}  & \multirow{2}{*}{Avg. ITA $\downarrow$}  & \multirow{2}{*}{Bias $\downarrow$} & \multirow{2}{*}{Score $\downarrow$} & \multirow{2}{*}{MAE $\downarrow$} & \multicolumn{6}{c}{ITA per skin type $\downarrow$} \\
\cline{6-11}
&  & & & & I & II & III & IV & V & VI \\
\midrule
\rowstyle{\color{dt}}  TRUST~\cite{feng2022towards} &   $13.87$ &   $\mathbf{2.79}$ &   $\mathbf{16.67}$ &   $\mathbf{18.41}$ &   $11.90$   & $11.87$   & $11.20$   & $13.92$  & $16.15$ & $18.21$ \\
\rowstyle{\color{dt}} ID2Albedo~\cite{ren2023improving} &  $\mathbf{12.07}$ & $4.91$ & $16.98$ & $23.33$ & $18.30$   & $9.13$   & $5.83$   & $9.46$  & $19.09$ & $\textbf{10.59}$ \\
\hline
Deep3D~\cite{deng2019accurate} & $22.57$ & $24.44$ & $47.02$   & $27.98$ & $\mathbf{8.92}$   & $\textbf{9.08}$   & $8.15$    & $10.90$   & $28.48$   & $69.90$ \\
GANFIT~\cite{Gecer19:ganfit} &  $62.29$ & $31.81$ & $94.11$   & $63.31$ & $94.80$   & $87.83$   & $76.25$   & $65.05$   & $38.24$   & $\textbf{11.59}$     \\
MGCNet~\cite{shang2020self} &  $21.41$ & $17.58$ & $38.99$   & $\mathbf{25.17}$  & $19.98$   & $12.76$   & $8.53$   & $\textbf{9.21}$   & $22.66$   & $55.34$    \\
DECA~\cite{Feng2021} &  $28.74$ & $29.24$ & $57.98$   & $38.17$  & $9.34$   & $11.66$   & $11.58$   & $16.69$   & $39.10$   & $84.06$   \\
INORig~\cite{Bai21} &  $27.68$ & $28.18$ & $55.86$   & $33.20$   & $23.25$   & $11.88$   & $\mathbf{4.86}$    & $9.75$    & $35.78$   & $80.54$    \\
CEST~\cite{Wen21}& $35.18$ & $12.14$ & $47.32$   & $29.92$  & $50.98$   & $38.77$   & $29.22$   & $23.62$   & $21.92$   & $46.57$     \\
\hline
Ours  &  \textbf{21.37} & \textbf{5.45} & \textbf{26.82}  & 28.09  &  31.09  & 19.48  & 14.76  &  16.96  & \textbf{20.52} & 25.42 \\
\bottomrule
\end{tabular}} %
\caption{Comparison to state-of-the-arts on the FAIR benchmark~\cite{feng2022towards}. We utilize the FAIR official metrics, such as average ITA error, bias score (standard deviation), total score (avg. ITA+Bias), mean average error, and average ITA score per skin type in degrees (I: very light, VI: very dark). Methods in grey font in the figure represent models trained using captured data.}
\label{tab:fair}
\end{table*}

Specifically, we flatten and concatenate all texture latents $\mathbf{z}_{uv \cdot 1}$, $\mathbf{z}_{uv \cdot 2}$, $\mathbf{z}_{uv \cdot 3}$, $\dots$, $\mathbf{z}_{uv \cdot n}$, denoted as $\mathbf{z}_{uv \cdot {all}} \in \mathbb{R}^{m \times d}$, where $m$ represents the total number of patches. We initialize a query latent $\mathbf{q} \in \mathbb{R}^{hw \times d}$ and define two different projectors $\mathcal{P}_{k}$ and $\mathcal{P}_{v}$ to obtain key latent $\mathbf{k}$ and value latent $\mathbf{v}$. Our cross-attention module computes the a fixed-length albedo latent $\mathbf{a} \in \mathbb{R}^{hw \times d}$ through $ \mathbf{a} = \text{Attention}(\mathbf{q}, \mathbf{k}, \mathbf{v}) = \text{softmax}\left(\mathbf{q}\mathbf{k}^T\right) \mathbf{v}$.
In this way, HiFiAlbedo can handle variable-length face input, allowing us to use a large number of face images for model training. During the testing phase, any number of images can be used for inference, even a single image. After obtaining the albedo latent $\mathbf{a}$, it is quantized and decoded by a pre-trained decoder $\mathcal{G}$ to produce the final albedo map.

To train our projector, query latent and cross-attention module in HiFiAlbedo, we use off-the-shelf  HRN~\cite{lei2023hierarchical} model to predict face shape from the input image and reuse its light estimation network for initialization.
To ensure the rendered images close to input faces, we continuously update the parameter of the light estimation network during the training process.
After differentiable rendering, we use L1 loss $\mathcal{L}_{rec}$ and LPIPS loss $\mathcal{L}_{lpips}$ to measure the error between rendered and real faces of the same identity. In addition, we propose a novel group identity loss $\mathcal{L}_{gid}$ to encourage the model to capture the overall appearance among multiple faces.
Specifically, our group identity loss is expressed as
\begin{equation}
\label{eq:grouploss}
\mathcal{L}_{gid} =  \frac{1}{n^2}\sum_{i=1}^{n} \sum_{j=1}^{n} w_{ij}\cdot(1 - \frac{\mathcal{A}(\mathbf{R}_{i})\mathcal{A}(\mathbf{I}_{j})}{\|\mathcal{A}(\mathbf{R}_{i})\|_2\cdot \|\mathcal{A}(\mathbf{I}_{j})\|_2}),
\end{equation}
where $w_{ij}$ represent the group weight calculated by two input faces $\mathbf{I}_i$ and $\mathbf{I}_j$, and $\mathbf{R}$ represents the warped albedo conditioned by face shape and illumination.
The intuition for using group weights $w_{ij}$ is that images of the same person inherently exhibit varying levels of similarity. Hence, group weights are employed to enhance the aggregation of identities within a batch. The details are as follows:
\begin{equation}
\label{eq:weight}
w_{ij} = \frac{\mathcal{A}(\mathbf{I}_{i})\mathcal{A}(\mathbf{I}_{j})}{\|\mathcal{A}(\mathbf{I}_{i})\|_2\cdot \|\mathcal{A}(\mathbf{I}_{j})\|_2}.
\end{equation}
In this way, we extend the original identity loss from a single image to the entire batch to constrain consistent albedo recovery.
Overall, our training loss function is defined as follows:
\begin{equation}
\label{eq:all3}
\mathcal{L}_{albedo} = \mathcal{L}_{gid} + \eta _{1}\mathcal{L}_{rec} + \eta _{2}\mathcal{L}_{lpips},
\end{equation}
where the loss weights $\eta_{1}$ and $\eta_{2}$ are set as $10$ and $0.1$, respectively.

\section{Experiments}
\subsection{Implementation Details}
\noindent{\bf Facial Texture Codebook Learning.} 
To improve the fidelity of the texture codebook, we downloaded high-resolution images from the WebFace260M dataset~\cite{zhu2021webface260m}. We collated 0.5 million images to ensure that each image contains a face with a resolution higher than $1024\times1024$.
These faces are normalized by the ArcFace template ~\cite{deng2019arcface,deng2020retinaface} at $1024\times1024$ resolution
to train a 10-epoch VQGAN~\cite{esser2021taming} model on 32$\times$A100 GPUs.

\noindent{\bf UV Texture Reconstruction.} During this phase, all the parameters of the previously trained VQGAN model were frozen to preserve its initial configuration. In addition, we trained an additional component, the face2texture encoder $\mathcal{E}_{uv}$, which is initialized by $\mathcal{E}$.  We pre-generated 200 UV textures from the CelebAHQ dataset using HRN~\cite{lei2023hierarchical}. We randomly select samples for adversarial training to correctly reconstruct the UV topology. The encoder $\mathcal{E}_{uv}$ is trained for 10 epochs on 32$\times$A100 GPUs.

\noindent{\bf High-Fidelity Albedo Disambiguation.} To ensure that the query latent perceives uniformly distributed light and removes occlusions from the albedo map, training the module requires the collection of multiple images of one subject in different scenes. For this purpose, we chose the VGGFace2~\cite{cao2018vggface2}, a widely-used facial recognition training set that is publicly accessible and contains 3.31 million images from 9131 subjects. In the training process, we set each subject to learn albedo from $n=4$ images simultaneously to balance training speed and memory consumption.  

\noindent{\bf Training Settings.} For all stages of training, we utilized the Adam optimizer with $\beta_{1}=0.9$ and $\beta_{2}=0.999$. We maintained a batch size of 4 per GPU, with a learning rate of 0.00001 for all stages. To improve training efficiency, we adopted mixed-precision training techniques~\cite{micikevicius2018mixed} and static graph compilation. Our training framework is PyTorch, complemented by PyTorch3D~\cite{ravi2020pytorch3d} for differentiable rendering.

\subsection{FAIR Benchmark Results}
\label{sec:exp:fair}
Following TRUST~\cite{feng2022towards} and ID2Albedo~\cite{ren2023improving}, we perform a qualitative and quantitative evaluation on the FAIR benchmark, shown in Fig.~\ref{fig:fairness} and Tab.~\ref{tab:fair}, respectively. 
In Fig.~\ref{fig:fairness}, there are two common problems with current methods, one is the strong bias towards certain skin color  types~\cite{Gecer19:ganfit,Bai21}, and the other is that existing unbiased albedo models have low fidelity~\cite{feng2022towards,ren2023improving}.
TRUST~\cite{feng2022towards}, ID2Albedo~\cite{ren2023improving}, and our method all perform albedo estimation well, but our method clearly achieves high fidelity results and faithfully reproduces the input face features.
Numerically, our method is not as good as the latest two methods that use captured data, but it is still much better than other methods that only use in-the-wild images.
Since our model is trained using large in-the-wild images where the skin color distribution is not uniform, the model performs worse on skin color types at both ends (Type I and VI). Overall, our method is close to existing captured data based methods in terms of bias score, but with higher visual fidelity.

\subsection{Real-World Results}
\label{sec:exp:real.results}

\begin{figure}[t]
\centering
\includegraphics[width=0.5\textwidth]{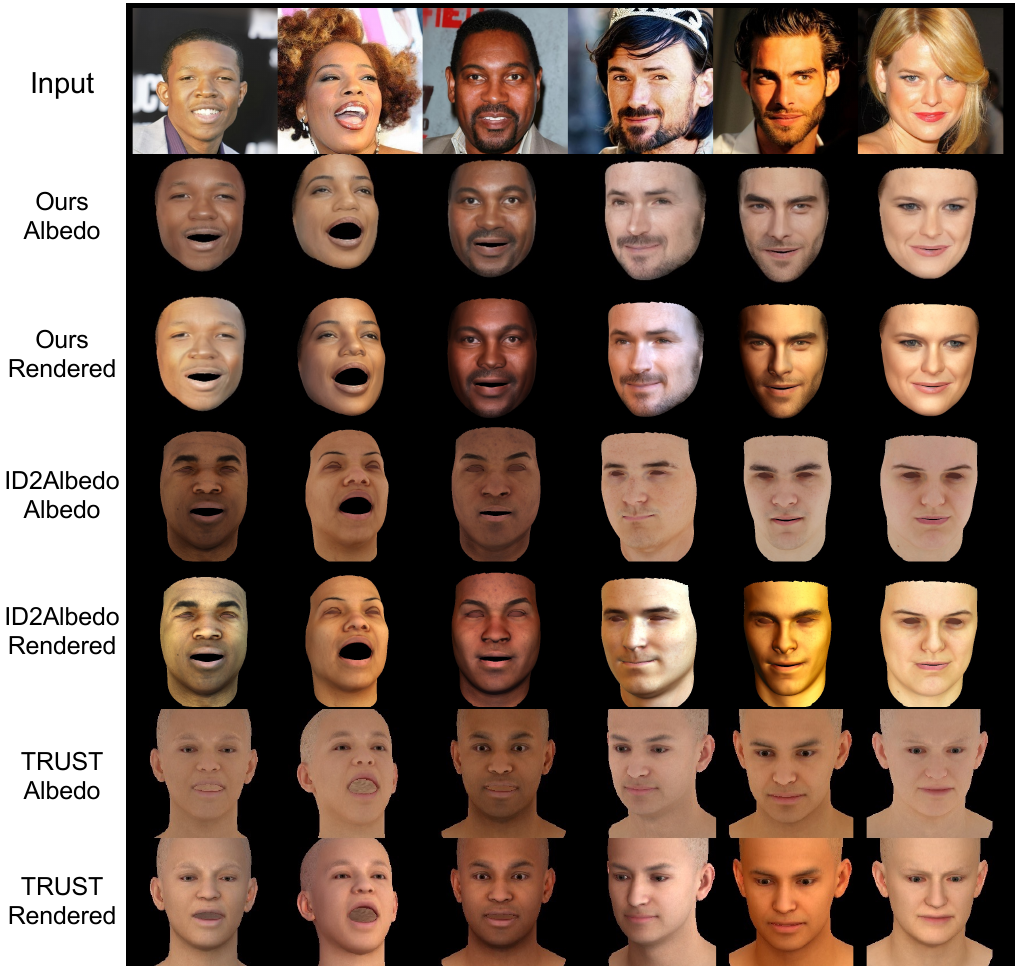}
\caption{Comparisons on in-the-wild images.  From top to bottom: inputs, ours, ID2Albedo~\cite{ren2023improving} and TRUST~\cite{feng2022towards}
albedo and rendered images.  We achieve the most realistic rendered results.}
\label{fig:real}
\end{figure}

To evaluate the robustness of our approach in real-world images, we qualitatively compare it with other methods on the CelebA-HQ dataset, as shown in Fig.~\ref{fig:real}. The results show that our albedo achieves the highest fidelity results while maintaining fairness.
We also analyze the quantitative results in CelebA-HQ. The results in Tab.~\ref{tab:real} show that our method achieves the best results in all metrics.

\begin{figure}[t]
    \centering
    \includegraphics[width=0.8\linewidth]{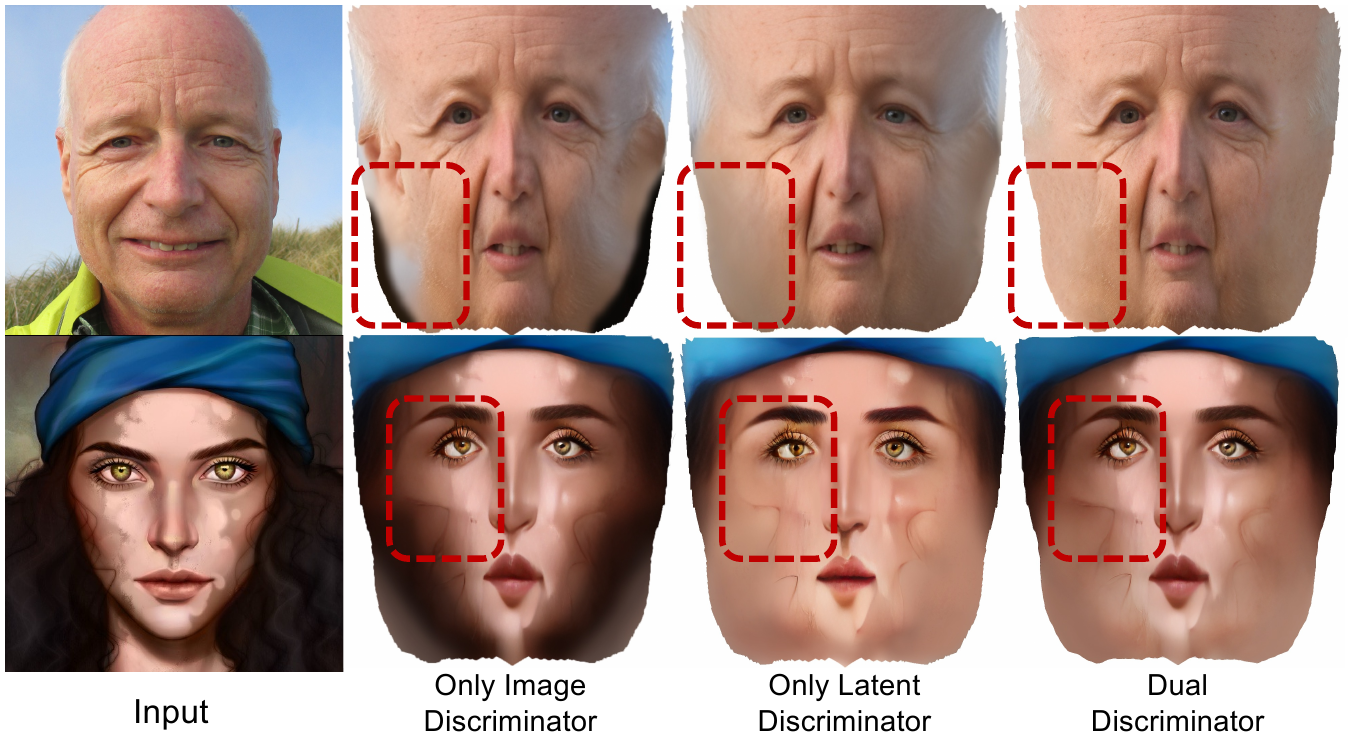}
    \caption{Ablation study of our dual discriminator. By adversarial supervision in latent and image spaces, we achieved high-quality UV texture unwrapping.} 
    \label{fig:dualdis} 
\end{figure}

\begin{figure}[t]
\centering
\includegraphics[width=\linewidth]{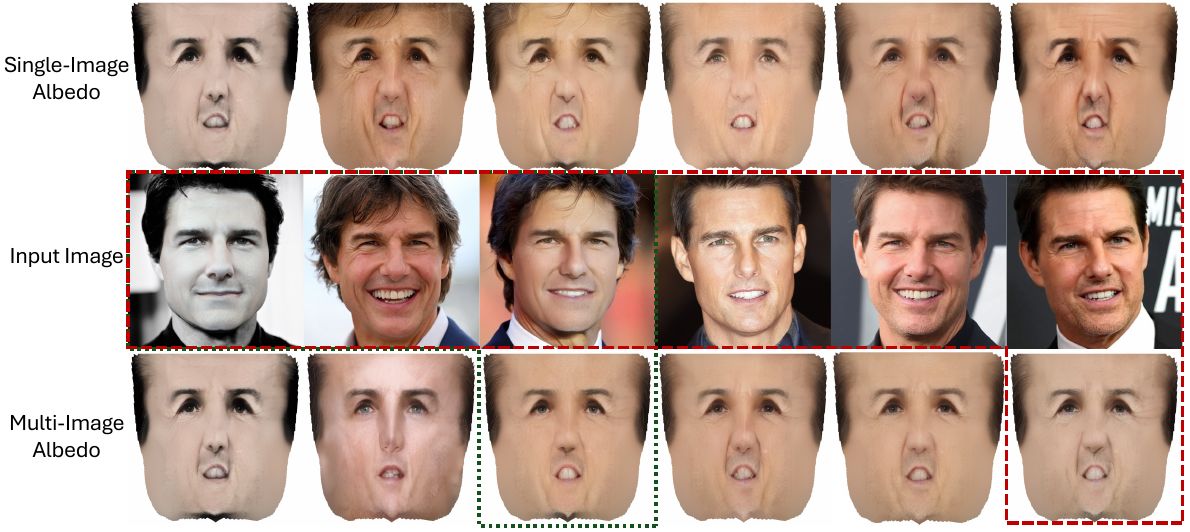}
\caption{Ablation study of multi-image inference. The first row displays single input image albedo maps, while the last row shows increasing multi input images generated albedo.}
\label{fig:latent_attention}
\end{figure}

\subsection{Ablation Studies}

\noindent{\bf Dual Discriminator.}
We first verify our dual discriminator proposed in Sec~\ref{Sec:Texture}. We train this module under different configurations and the results are shown in Fig~\ref{fig:dualdis}.
We observe that using the image discriminator alone leads to UV artifacts (See the black shadows on the edge parts of the first row), while using the latent space alone does not lead to artifacts, but loses facial details (See the texture details on the cheeks in the second row). In contrast, using both methods together results in high quality UV topography while preserving details.

\begin{table}[t]
\centering
\resizebox{0.9\linewidth}{!}{%
\begin{tabular}{l|cccc}
\toprule
Method & PSNR$\uparrow$ & SSIM$\uparrow$ & LPIPS$\downarrow$ & ID$\uparrow$ \\ 
\midrule
TRUST~\cite{feng2022towards} & 21.63 & 0.8520 & 0.2014 & 0.4780 \\ 
ID2Albedo~\cite{ren2023improving} & 23.72 & 0.8840 & 0.1549 & 0.5320 \\
Ours & \textbf{27.88} & \textbf{0.9060} & \textbf{0.1413} & \textbf{0.5870} \\
\bottomrule
\end{tabular}}
\caption{Comparisons with previous albedo estimation methods. All metrics are computed on the CelebAMask-HQ dataset~\cite{CelebAMask-HQ}.}
\label{tab:real}
\end{table}

\begin{table}[t]
\centering
\resizebox{0.9\linewidth}{!}{%
\begin{tabular}{l|cccc}
\toprule
Method & PSNR$\uparrow$ & SSIM$\uparrow$ & LPIPS$\downarrow$ & ID$\uparrow$ \\ 
\midrule
Baseline              & 24.55	&	0.8737 	&	0.1514	&	0.5076 	\\
+Sampling Constraint  & 25.41	&	0.8913 	&	0.1447	&	0.5215 	\\
+Group Identity Loss   & 26.54	&	0.8941 	&	0.1425	&	0.5672 	\\
Both                 & \textbf{27.88} & \textbf{0.9060} & \textbf{0.1413} & \textbf{0.5870} \\
\bottomrule
\end{tabular}}
\caption{Ablation study of sampling constraint and the group identity loss. All metrics are computed on the CelebAMask-HQ dataset~\cite{CelebAMask-HQ}.}
\label{tab:abl:group-wise-arcface-loss}
\end{table}

\noindent{\bf Multi-image Inference.}
We investigate the advantages of cross-attention based multi-image inference. As can be seen in the first row of Fig.~\ref{fig:latent_attention}, model inference for a single image suffers from albedo/illumination ambiguity, whereas the quality of the albedo inference for multiple images improves with the number of input images (see the last row).
For example, the green box represents the albedo map obtained from the first three inputs, while the red box corresponds to the map derived from the all six inputs.
However, the number of input images does not have to increase all the time, we find that the albedo stabilizes when the number of input images is greater than or equal to three.

\noindent{\bf Group Identity Loss.}
We finally validate our sampling constraint and group identity loss. As shown in Tab.~\ref{tab:abl:group-wise-arcface-loss}, the sampling constraint yields better inputs, which improves the consistency of albedo recovery, leading to improved performance. The group identity loss, on the other hand, guides the model updates via better optimization direction, further improving the reconstruction quality. Our HiFiAlbedo achieves the best results by using both modules.

\section{Conclusion}
In this paper, we introduce HiFiAlbedo, a novel model for high-fidelity albedo reconstruction. Our model doesn't require captured data and can be implemented with only in-the-wild images, significantly reducing data costs. HiFiAlbedo uses a vector-quantized generative model and a cross-attention module to infer unbiased albedo from multiple facial images of the same individual. Experiments demonstrate that our proposed method achieves competitive performance on the FAIR benchmark and excellent generalizability and fairness on real-world images. Our method can not only be used for high-quality rendering, but also opens a new avenue for albedo recovery from facial images.

\paragraph{Acknowledgements} L.Z was supported in part by General Program of National Natural Science Foundation of China (62372403).

\bibliographystyle{ACM-Reference-Format}
\bibliography{siggraph_aisa}


\begin{thebibliography}{44}


\ifx \showCODEN    \undefined \def \showCODEN     #1{\unskip}     \fi
\ifx \showDOI      \undefined \def \showDOI       #1{#1}\fi
\ifx \showISBNx    \undefined \def \showISBNx     #1{\unskip}     \fi
\ifx \showISBNxiii \undefined \def \showISBNxiii  #1{\unskip}     \fi
\ifx \showISSN     \undefined \def \showISSN      #1{\unskip}     \fi
\ifx \showLCCN     \undefined \def \showLCCN      #1{\unskip}     \fi
\ifx \shownote     \undefined \def \shownote      #1{#1}          \fi
\ifx \showarticletitle \undefined \def \showarticletitle #1{#1}   \fi
\ifx \showURL      \undefined \def \showURL       {\relax}        \fi
\providecommand\bibfield[2]{#2}
\providecommand\bibinfo[2]{#2}
\providecommand\natexlab[1]{#1}
\providecommand\showeprint[2][]{arXiv:#2}

\bibitem[Aldrian and Smith(2012)]%
        {Aldrian12}
\bibfield{author}{\bibinfo{person}{Oswald Aldrian} {and}
  \bibinfo{person}{William~AP Smith}.} \bibinfo{year}{2012}\natexlab{}.
\newblock \showarticletitle{Inverse rendering of faces with a 3D morphable
  model}.
\newblock \bibinfo{journal}{\emph{IEEE transactions on pattern analysis and
  machine intelligence}} \bibinfo{volume}{35}, \bibinfo{number}{5}
  (\bibinfo{year}{2012}), \bibinfo{pages}{1080--1093}.
\newblock


\bibitem[Bai et~al\mbox{.}(2023)]%
        {bai2023ffhq}
\bibfield{author}{\bibinfo{person}{Haoran Bai}, \bibinfo{person}{Di Kang},
  \bibinfo{person}{Haoxian Zhang}, \bibinfo{person}{Jinshan Pan}, {and}
  \bibinfo{person}{Linchao Bao}.} \bibinfo{year}{2023}\natexlab{}.
\newblock \showarticletitle{Ffhq-uv: Normalized facial uv-texture dataset for
  3d face reconstruction}. In \bibinfo{booktitle}{\emph{Proceedings of the
  IEEE/CVF Conference on Computer Vision and Pattern Recognition}}.
  \bibinfo{pages}{362--371}.
\newblock


\bibitem[Bai et~al\mbox{.}(2021)]%
        {Bai21}
\bibfield{author}{\bibinfo{person}{Ziqian Bai}, \bibinfo{person}{Zhaopeng Cui},
  \bibinfo{person}{Xiaoming Liu}, {and} \bibinfo{person}{Ping Tan}.}
  \bibinfo{year}{2021}\natexlab{}.
\newblock \showarticletitle{Riggable 3d face reconstruction via in-network
  optimization}. In \bibinfo{booktitle}{\emph{Proceedings of the IEEE/CVF
  conference on computer vision and pattern recognition}}.
  \bibinfo{pages}{6216--6225}.
\newblock


\bibitem[Cao et~al\mbox{.}(2018)]%
        {cao2018vggface2}
\bibfield{author}{\bibinfo{person}{Qiong Cao}, \bibinfo{person}{Li Shen},
  \bibinfo{person}{Weidi Xie}, \bibinfo{person}{Omkar~M Parkhi}, {and}
  \bibinfo{person}{Andrew Zisserman}.} \bibinfo{year}{2018}\natexlab{}.
\newblock \showarticletitle{Vggface2: A dataset for recognising faces across
  pose and age}. In \bibinfo{booktitle}{\emph{2018 13th IEEE international
  conference on automatic face \& gesture recognition (FG 2018)}}. IEEE,
  \bibinfo{pages}{67--74}.
\newblock


\bibitem[Debevec et~al\mbox{.}(2012)]%
        {debevec2012light}
\bibfield{author}{\bibinfo{person}{Paul Debevec}, \bibinfo{person}{Paul
  Graham}, \bibinfo{person}{Jay Busch}, {and} \bibinfo{person}{Mark Bolas}.}
  \bibinfo{year}{2012}\natexlab{}.
\newblock \showarticletitle{A single-shot light probe}.
\newblock In \bibinfo{booktitle}{\emph{ACM SIGGRAPH 2012 Talks}}.
  \bibinfo{pages}{1--1}.
\newblock


\bibitem[Deng et~al\mbox{.}(2018)]%
        {deng2018uv}
\bibfield{author}{\bibinfo{person}{Jiankang Deng}, \bibinfo{person}{Shiyang
  Cheng}, \bibinfo{person}{Niannan Xue}, \bibinfo{person}{Yuxiang Zhou}, {and}
  \bibinfo{person}{Stefanos Zafeiriou}.} \bibinfo{year}{2018}\natexlab{}.
\newblock \showarticletitle{Uv-gan: Adversarial facial uv map completion for
  pose-invariant face recognition}. In \bibinfo{booktitle}{\emph{Proceedings of
  the IEEE conference on computer vision and pattern recognition}}.
  \bibinfo{pages}{7093--7102}.
\newblock


\bibitem[Deng et~al\mbox{.}(2020)]%
        {deng2020retinaface}
\bibfield{author}{\bibinfo{person}{Jiankang Deng}, \bibinfo{person}{Jia Guo},
  \bibinfo{person}{Evangelos Ververas}, \bibinfo{person}{Irene Kotsia}, {and}
  \bibinfo{person}{Stefanos Zafeiriou}.} \bibinfo{year}{2020}\natexlab{}.
\newblock \showarticletitle{Retinaface: Single-shot multi-level face
  localisation in the wild}. In \bibinfo{booktitle}{\emph{Proceedings of the
  IEEE/CVF conference on computer vision and pattern recognition}}.
  \bibinfo{pages}{5203--5212}.
\newblock


\bibitem[Deng et~al\mbox{.}(2019a)]%
        {deng2019arcface}
\bibfield{author}{\bibinfo{person}{Jiankang Deng}, \bibinfo{person}{Jia Guo},
  \bibinfo{person}{Niannan Xue}, {and} \bibinfo{person}{Stefanos Zafeiriou}.}
  \bibinfo{year}{2019}\natexlab{a}.
\newblock \showarticletitle{Arcface: Additive angular margin loss for deep face
  recognition}. In \bibinfo{booktitle}{\emph{Proceedings of the IEEE/CVF
  conference on computer vision and pattern recognition}}.
  \bibinfo{pages}{4690--4699}.
\newblock


\bibitem[Deng et~al\mbox{.}(2019b)]%
        {deng2019accurate}
\bibfield{author}{\bibinfo{person}{Yu Deng}, \bibinfo{person}{Jiaolong Yang},
  \bibinfo{person}{Sicheng Xu}, \bibinfo{person}{Dong Chen},
  \bibinfo{person}{Yunde Jia}, {and} \bibinfo{person}{Xin Tong}.}
  \bibinfo{year}{2019}\natexlab{b}.
\newblock \showarticletitle{Accurate 3d face reconstruction with
  weakly-supervised learning: From single image to image set}. In
  \bibinfo{booktitle}{\emph{Proceedings of the IEEE/CVF conference on computer
  vision and pattern recognition workshops}}. \bibinfo{pages}{0--0}.
\newblock


\bibitem[Egger et~al\mbox{.}(2018)]%
        {Egger18}
\bibfield{author}{\bibinfo{person}{Bernhard Egger}, \bibinfo{person}{Sandro
  Sch{\"o}nborn}, \bibinfo{person}{Andreas Schneider}, \bibinfo{person}{Adam
  Kortylewski}, \bibinfo{person}{Andreas Morel-Forster},
  \bibinfo{person}{Clemens Blumer}, {and} \bibinfo{person}{Thomas Vetter}.}
  \bibinfo{year}{2018}\natexlab{}.
\newblock \showarticletitle{Occlusion-aware 3d morphable models and an
  illumination prior for face image analysis}.
\newblock \bibinfo{journal}{\emph{International Journal of Computer Vision}}
  \bibinfo{volume}{126} (\bibinfo{year}{2018}), \bibinfo{pages}{1269--1287}.
\newblock


\bibitem[Egger et~al\mbox{.}(2020)]%
        {Egger2020}
\bibfield{author}{\bibinfo{person}{Bernhard Egger}, \bibinfo{person}{William~AP
  Smith}, \bibinfo{person}{Ayush Tewari}, \bibinfo{person}{Stefanie Wuhrer},
  \bibinfo{person}{Michael Zollhoefer}, \bibinfo{person}{Thabo Beeler},
  \bibinfo{person}{Florian Bernard}, \bibinfo{person}{Timo Bolkart},
  \bibinfo{person}{Adam Kortylewski}, \bibinfo{person}{Sami Romdhani},
  {et~al\mbox{.}}} \bibinfo{year}{2020}\natexlab{}.
\newblock \showarticletitle{3d morphable face models—past, present, and
  future}.
\newblock \bibinfo{journal}{\emph{ACM Transactions on Graphics (ToG)}}
  \bibinfo{volume}{39}, \bibinfo{number}{5} (\bibinfo{year}{2020}),
  \bibinfo{pages}{1--38}.
\newblock


\bibitem[Esser et~al\mbox{.}(2021)]%
        {esser2021taming}
\bibfield{author}{\bibinfo{person}{Patrick Esser}, \bibinfo{person}{Robin
  Rombach}, {and} \bibinfo{person}{Bjorn Ommer}.}
  \bibinfo{year}{2021}\natexlab{}.
\newblock \showarticletitle{Taming transformers for high-resolution image
  synthesis}. In \bibinfo{booktitle}{\emph{Proceedings of the IEEE/CVF
  conference on computer vision and pattern recognition}}.
  \bibinfo{pages}{12873--12883}.
\newblock


\bibitem[Feng et~al\mbox{.}(2022)]%
        {feng2022towards}
\bibfield{author}{\bibinfo{person}{Haiwen Feng}, \bibinfo{person}{Timo
  Bolkart}, \bibinfo{person}{Joachim Tesch}, \bibinfo{person}{Michael~J Black},
  {and} \bibinfo{person}{Victoria Abrevaya}.} \bibinfo{year}{2022}\natexlab{}.
\newblock \showarticletitle{Towards racially unbiased skin tone estimation via
  scene disambiguation}. In \bibinfo{booktitle}{\emph{European Conference on
  Computer Vision}}. Springer, \bibinfo{pages}{72--90}.
\newblock


\bibitem[Feng et~al\mbox{.}(2021)]%
        {Feng2021}
\bibfield{author}{\bibinfo{person}{Yao Feng}, \bibinfo{person}{Haiwen Feng},
  \bibinfo{person}{Michael~J Black}, {and} \bibinfo{person}{Timo Bolkart}.}
  \bibinfo{year}{2021}\natexlab{}.
\newblock \showarticletitle{Learning an animatable detailed 3D face model from
  in-the-wild images}.
\newblock \bibinfo{journal}{\emph{ACM Transactions on Graphics (ToG)}}
  \bibinfo{volume}{40}, \bibinfo{number}{4} (\bibinfo{year}{2021}),
  \bibinfo{pages}{1--13}.
\newblock


\bibitem[Gecer et~al\mbox{.}(2019)]%
        {Gecer19:ganfit}
\bibfield{author}{\bibinfo{person}{Baris Gecer}, \bibinfo{person}{Stylianos
  Ploumpis}, \bibinfo{person}{Irene Kotsia}, {and} \bibinfo{person}{Stefanos
  Zafeiriou}.} \bibinfo{year}{2019}\natexlab{}.
\newblock \showarticletitle{Ganfit: Generative adversarial network fitting for
  high fidelity 3d face reconstruction}. In
  \bibinfo{booktitle}{\emph{Proceedings of the IEEE/CVF conference on computer
  vision and pattern recognition}}.
\newblock


\bibitem[Gecer et~al\mbox{.}(2021)]%
        {gecer2021fast}
\bibfield{author}{\bibinfo{person}{Baris Gecer}, \bibinfo{person}{Stylianos
  Ploumpis}, \bibinfo{person}{Irene Kotsia}, {and} \bibinfo{person}{Stefanos~P
  Zafeiriou}.} \bibinfo{year}{2021}\natexlab{}.
\newblock \showarticletitle{Fast-GANFIT: Generative Adversarial Network for
  High Fidelity 3D Face Reconstruction}.
\newblock \bibinfo{journal}{\emph{IEEE TPAMI}} (\bibinfo{year}{2021}).
\newblock


\bibitem[Ghosh et~al\mbox{.}(2011)]%
        {ghosh2011multiview}
\bibfield{author}{\bibinfo{person}{Abhijeet Ghosh}, \bibinfo{person}{Graham
  Fyffe}, \bibinfo{person}{Borom Tunwattanapong}, \bibinfo{person}{Jay Busch},
  \bibinfo{person}{Xueming Yu}, {and} \bibinfo{person}{Paul Debevec}.}
  \bibinfo{year}{2011}\natexlab{}.
\newblock \showarticletitle{Multiview face capture using polarized spherical
  gradient illumination}. In \bibinfo{booktitle}{\emph{Proceedings of the 2011
  SIGGRAPH Asia Conference}}. \bibinfo{pages}{1--10}.
\newblock


\bibitem[Guo and Deng(2019)]%
        {insightface2019}
\bibfield{author}{\bibinfo{person}{Jia Guo} {and} \bibinfo{person}{Jiankang
  Deng}.} \bibinfo{year}{2019}\natexlab{}.
\newblock \bibinfo{title}{InsightFace Repo}.
\newblock
  \bibinfo{howpublished}{\url{https://github.com/deepinsight/insightface}}.
\newblock


\bibitem[Hu et~al\mbox{.}(2013)]%
        {Hu13}
\bibfield{author}{\bibinfo{person}{Guosheng Hu}, \bibinfo{person}{Pouria
  Mortazavian}, \bibinfo{person}{Josef Kittler}, {and} \bibinfo{person}{William
  Christmas}.} \bibinfo{year}{2013}\natexlab{}.
\newblock \showarticletitle{A facial symmetry prior for improved illumination
  fitting of 3D morphable model}. In \bibinfo{booktitle}{\emph{2013
  International Conference on Biometrics (ICB)}}. IEEE, \bibinfo{pages}{1--6}.
\newblock


\bibitem[Karras et~al\mbox{.}(2019)]%
        {karras2019style}
\bibfield{author}{\bibinfo{person}{Tero Karras}, \bibinfo{person}{Samuli
  Laine}, {and} \bibinfo{person}{Timo Aila}.} \bibinfo{year}{2019}\natexlab{}.
\newblock \showarticletitle{A style-based generator architecture for generative
  adversarial networks}. In \bibinfo{booktitle}{\emph{Proceedings of the
  IEEE/CVF conference on computer vision and pattern recognition}}.
  \bibinfo{pages}{4401--4410}.
\newblock


\bibitem[Khakhulin et~al\mbox{.}(2022)]%
        {Khakhulin2022realistic}
\bibfield{author}{\bibinfo{person}{Taras Khakhulin}, \bibinfo{person}{Vanessa
  Sklyarova}, \bibinfo{person}{Victor Lempitsky}, {and} \bibinfo{person}{Egor
  Zakharov}.} \bibinfo{year}{2022}\natexlab{}.
\newblock \showarticletitle{Realistic one-shot mesh-based head avatars}. In
  \bibinfo{booktitle}{\emph{European Conference on Computer Vision}}. Springer,
  \bibinfo{pages}{345--362}.
\newblock


\bibitem[Lattas et~al\mbox{.}(2020)]%
        {Lattas20}
\bibfield{author}{\bibinfo{person}{Alexandros Lattas},
  \bibinfo{person}{Stylianos Moschoglou}, \bibinfo{person}{Baris Gecer},
  \bibinfo{person}{Stylianos Ploumpis}, \bibinfo{person}{Vasileios
  Triantafyllou}, \bibinfo{person}{Abhijeet Ghosh}, {and}
  \bibinfo{person}{Stefanos Zafeiriou}.} \bibinfo{year}{2020}\natexlab{}.
\newblock \showarticletitle{AvatarMe: Realistically Renderable 3D Facial
  Reconstruction" in-the-wild"}. In \bibinfo{booktitle}{\emph{Proceedings of
  the IEEE/CVF conference on computer vision and pattern recognition}}.
  \bibinfo{pages}{760--769}.
\newblock


\bibitem[Lattas et~al\mbox{.}(2023)]%
        {lattas2023fitme}
\bibfield{author}{\bibinfo{person}{Alexandros Lattas},
  \bibinfo{person}{Stylianos Moschoglou}, \bibinfo{person}{Stylianos Ploumpis},
  \bibinfo{person}{Baris Gecer}, \bibinfo{person}{Jiankang Deng}, {and}
  \bibinfo{person}{Stefanos Zafeiriou}.} \bibinfo{year}{2023}\natexlab{}.
\newblock \showarticletitle{Fitme: Deep photorealistic 3d morphable model
  avatars}. In \bibinfo{booktitle}{\emph{Proceedings of the IEEE/CVF Conference
  on Computer Vision and Pattern Recognition}}. \bibinfo{pages}{8629--8640}.
\newblock


\bibitem[Lattas et~al\mbox{.}(2021)]%
        {lattas2021avatarme++}
\bibfield{author}{\bibinfo{person}{Alexandros Lattas},
  \bibinfo{person}{Stylianos Moschoglou}, \bibinfo{person}{Stylianos Ploumpis},
  \bibinfo{person}{Baris Gecer}, \bibinfo{person}{Abhijeet Ghosh}, {and}
  \bibinfo{person}{Stefanos Zafeiriou}.} \bibinfo{year}{2021}\natexlab{}.
\newblock \showarticletitle{Avatarme++: Facial shape and brdf inference with
  photorealistic rendering-aware gans}.
\newblock \bibinfo{journal}{\emph{IEEE Transactions on Pattern Analysis and
  Machine Intelligence}} \bibinfo{volume}{44}, \bibinfo{number}{12}
  (\bibinfo{year}{2021}), \bibinfo{pages}{9269--9284}.
\newblock


\bibitem[Lee et~al\mbox{.}(2020)]%
        {CelebAMask-HQ}
\bibfield{author}{\bibinfo{person}{Cheng-Han Lee}, \bibinfo{person}{Ziwei Liu},
  \bibinfo{person}{Lingyun Wu}, {and} \bibinfo{person}{Ping Luo}.}
  \bibinfo{year}{2020}\natexlab{}.
\newblock \showarticletitle{Maskgan: Towards diverse and interactive facial
  image manipulation}. In \bibinfo{booktitle}{\emph{Proceedings of the IEEE/CVF
  conference on computer vision and pattern recognition}}.
  \bibinfo{pages}{5549--5558}.
\newblock


\bibitem[LeGendre et~al\mbox{.}(2018)]%
        {LeGendre2018facial}
\bibfield{author}{\bibinfo{person}{Chloe LeGendre}, \bibinfo{person}{Kalle
  Bladin}, \bibinfo{person}{Bipin Kishore}, \bibinfo{person}{Xinglei Ren},
  \bibinfo{person}{Xueming Yu}, {and} \bibinfo{person}{Paul Debevec}.}
  \bibinfo{year}{2018}\natexlab{}.
\newblock \showarticletitle{Efficient multispectral facial capture with
  monochrome cameras}.
\newblock In \bibinfo{booktitle}{\emph{ACM SIGGRAPH 2018 Posters}}.
  \bibinfo{pages}{1--2}.
\newblock


\bibitem[Lei et~al\mbox{.}(2023)]%
        {lei2023hierarchical}
\bibfield{author}{\bibinfo{person}{Biwen Lei}, \bibinfo{person}{Jianqiang Ren},
  \bibinfo{person}{Mengyang Feng}, \bibinfo{person}{Miaomiao Cui}, {and}
  \bibinfo{person}{Xuansong Xie}.} \bibinfo{year}{2023}\natexlab{}.
\newblock \showarticletitle{A hierarchical representation network for accurate
  and detailed face reconstruction from in-the-wild images}. In
  \bibinfo{booktitle}{\emph{Proceedings of the IEEE/CVF Conference on Computer
  Vision and Pattern Recognition}}. \bibinfo{pages}{394--403}.
\newblock


\bibitem[Micikevicius et~al\mbox{.}(2018)]%
        {micikevicius2018mixed}
\bibfield{author}{\bibinfo{person}{Paulius Micikevicius},
  \bibinfo{person}{Sharan Narang}, \bibinfo{person}{Jonah Alben},
  \bibinfo{person}{Gregory Diamos}, \bibinfo{person}{Erich Elsen},
  \bibinfo{person}{David Garcia}, \bibinfo{person}{Boris Ginsburg},
  \bibinfo{person}{Michael Houston}, \bibinfo{person}{Oleksii Kuchaiev},
  \bibinfo{person}{Ganesh Venkatesh}, {and} \bibinfo{person}{Hao Wu}.}
  \bibinfo{year}{2018}\natexlab{}.
\newblock \showarticletitle{Mixed Precision Training}. In
  \bibinfo{booktitle}{\emph{ICLR}}.
\newblock


\bibitem[Papantoniou et~al\mbox{.}(2023)]%
        {papantoniou2023relightify}
\bibfield{author}{\bibinfo{person}{Foivos~Paraperas Papantoniou},
  \bibinfo{person}{Alexandros Lattas}, \bibinfo{person}{Stylianos Moschoglou},
  {and} \bibinfo{person}{Stefanos Zafeiriou}.} \bibinfo{year}{2023}\natexlab{}.
\newblock \showarticletitle{Relightify: Relightable 3d faces from a single
  image via diffusion models}. In \bibinfo{booktitle}{\emph{Proceedings of the
  IEEE/CVF International Conference on Computer Vision}}.
  \bibinfo{pages}{8806--8817}.
\newblock


\bibitem[Paysan et~al\mbox{.}(2009)]%
        {Paysan09}
\bibfield{author}{\bibinfo{person}{Pascal Paysan}, \bibinfo{person}{Reinhard
  Knothe}, \bibinfo{person}{Brian Amberg}, \bibinfo{person}{Sami Romdhani},
  {and} \bibinfo{person}{Thomas Vetter}.} \bibinfo{year}{2009}\natexlab{}.
\newblock \showarticletitle{A 3D face model for pose and illumination invariant
  face recognition}. In \bibinfo{booktitle}{\emph{2009 sixth IEEE international
  conference on advanced video and signal based surveillance}}.
\newblock


\bibitem[Ramamoorthi and Hanrahan(2001)]%
        {Ramamoorthi01}
\bibfield{author}{\bibinfo{person}{Ravi Ramamoorthi} {and} \bibinfo{person}{Pat
  Hanrahan}.} \bibinfo{year}{2001}\natexlab{}.
\newblock \showarticletitle{A signal-processing framework for inverse
  rendering}. In \bibinfo{booktitle}{\emph{Proceedings of the 28th annual
  conference on Computer graphics and interactive techniques}}.
  \bibinfo{pages}{117--128}.
\newblock


\bibitem[Ravi et~al\mbox{.}(2020)]%
        {ravi2020pytorch3d}
\bibfield{author}{\bibinfo{person}{Nikhila Ravi}, \bibinfo{person}{Jeremy
  Reizenstein}, \bibinfo{person}{David Novotny}, \bibinfo{person}{Taylor
  Gordon}, \bibinfo{person}{Wan-Yen Lo}, \bibinfo{person}{Justin Johnson},
  {and} \bibinfo{person}{Georgia Gkioxari}.} \bibinfo{year}{2020}\natexlab{}.
\newblock \showarticletitle{Accelerating 3d deep learning with pytorch3d}.
\newblock \bibinfo{journal}{\emph{arXiv preprint arXiv:2007.08501}}
  (\bibinfo{year}{2020}).
\newblock


\bibitem[Ren et~al\mbox{.}(2024)]%
        {ren2024monocular}
\bibfield{author}{\bibinfo{person}{Xingyu Ren}, \bibinfo{person}{Jiankang
  Deng}, \bibinfo{person}{Yuhao Cheng}, \bibinfo{person}{Jia Guo},
  \bibinfo{person}{Chao Ma}, \bibinfo{person}{Yichao Yan},
  \bibinfo{person}{Wenhan Zhu}, {and} \bibinfo{person}{Xiaokang Yang}.}
  \bibinfo{year}{2024}\natexlab{}.
\newblock \showarticletitle{Monocular Identity-Conditioned Facial Reflectance
  Reconstruction}. In \bibinfo{booktitle}{\emph{Proceedings of the IEEE/CVF
  Conference on Computer Vision and Pattern Recognition}}.
  \bibinfo{pages}{885--895}.
\newblock


\bibitem[Ren et~al\mbox{.}(2023a)]%
        {ren2023improving}
\bibfield{author}{\bibinfo{person}{Xingyu Ren}, \bibinfo{person}{Jiankang
  Deng}, \bibinfo{person}{Chao Ma}, \bibinfo{person}{Yichao Yan}, {and}
  \bibinfo{person}{Xiaokang Yang}.} \bibinfo{year}{2023}\natexlab{a}.
\newblock \showarticletitle{Improving Fairness in Facial Albedo Estimation via
  Visual-Textual Cues}. In \bibinfo{booktitle}{\emph{Proceedings of the
  IEEE/CVF Conference on Computer Vision and Pattern Recognition}}.
  \bibinfo{pages}{4511--4520}.
\newblock


\bibitem[Ren et~al\mbox{.}(2023b)]%
        {ren2023facial}
\bibfield{author}{\bibinfo{person}{Xingyu Ren}, \bibinfo{person}{Alexandros
  Lattas}, \bibinfo{person}{Baris Gecer}, \bibinfo{person}{Jiankang Deng},
  \bibinfo{person}{Chao Ma}, {and} \bibinfo{person}{Xiaokang Yang}.}
  \bibinfo{year}{2023}\natexlab{b}.
\newblock \showarticletitle{Facial geometric detail recovery via implicit
  representation}. In \bibinfo{booktitle}{\emph{2023 IEEE 17th International
  Conference on Automatic Face and Gesture Recognition (FG)}}.
\newblock


\bibitem[Rombach et~al\mbox{.}(2022)]%
        {rombach2022high}
\bibfield{author}{\bibinfo{person}{Robin Rombach}, \bibinfo{person}{Andreas
  Blattmann}, \bibinfo{person}{Dominik Lorenz}, \bibinfo{person}{Patrick
  Esser}, {and} \bibinfo{person}{Bj{\"o}rn Ommer}.}
  \bibinfo{year}{2022}\natexlab{}.
\newblock \showarticletitle{High-resolution image synthesis with latent
  diffusion models}. In \bibinfo{booktitle}{\emph{Proceedings of the IEEE/CVF
  conference on computer vision and pattern recognition}}.
  \bibinfo{pages}{10684--10695}.
\newblock


\bibitem[Shang et~al\mbox{.}(2020)]%
        {shang2020self}
\bibfield{author}{\bibinfo{person}{Jiaxiang Shang}, \bibinfo{person}{Tianwei
  Shen}, \bibinfo{person}{Shiwei Li}, \bibinfo{person}{Lei Zhou},
  \bibinfo{person}{Mingmin Zhen}, \bibinfo{person}{Tian Fang}, {and}
  \bibinfo{person}{Long Quan}.} \bibinfo{year}{2020}\natexlab{}.
\newblock \showarticletitle{Self-supervised monocular 3d face reconstruction by
  occlusion-aware multi-view geometry consistency}. In
  \bibinfo{booktitle}{\emph{European Conference on Computer Vision}}.
\newblock


\bibitem[Smith et~al\mbox{.}(2020)]%
        {smith2020morphable}
\bibfield{author}{\bibinfo{person}{William~AP Smith}, \bibinfo{person}{Alassane
  Seck}, \bibinfo{person}{Hannah Dee}, \bibinfo{person}{Bernard Tiddeman},
  \bibinfo{person}{Joshua~B Tenenbaum}, {and} \bibinfo{person}{Bernhard
  Egger}.} \bibinfo{year}{2020}\natexlab{}.
\newblock \showarticletitle{A morphable face albedo model}. In
  \bibinfo{booktitle}{\emph{Proceedings of the IEEE/CVF Conference on Computer
  Vision and Pattern Recognition}}. \bibinfo{pages}{5011--5020}.
\newblock


\bibitem[Tewari et~al\mbox{.}(2022)]%
        {tewari2022advances}
\bibfield{author}{\bibinfo{person}{Ayush Tewari}, \bibinfo{person}{Justus
  Thies}, \bibinfo{person}{Ben Mildenhall}, \bibinfo{person}{Pratul
  Srinivasan}, \bibinfo{person}{Edgar Tretschk}, \bibinfo{person}{Wang Yifan},
  \bibinfo{person}{Christoph Lassner}, \bibinfo{person}{Vincent Sitzmann},
  \bibinfo{person}{Ricardo Martin-Brualla}, \bibinfo{person}{Stephen Lombardi},
  {et~al\mbox{.}}} \bibinfo{year}{2022}\natexlab{}.
\newblock \showarticletitle{Advances in neural rendering}. In
  \bibinfo{booktitle}{\emph{Computer Graphics Forum}},
  Vol.~\bibinfo{volume}{41}. Wiley Online Library, \bibinfo{pages}{703--735}.
\newblock


\bibitem[Van Den~Oord et~al\mbox{.}(2017)]%
        {van2017neural}
\bibfield{author}{\bibinfo{person}{Aaron Van Den~Oord}, \bibinfo{person}{Oriol
  Vinyals}, {et~al\mbox{.}}} \bibinfo{year}{2017}\natexlab{}.
\newblock \showarticletitle{Neural discrete representation learning}.
\newblock \bibinfo{journal}{\emph{Advances in neural information processing
  systems}}  \bibinfo{volume}{30} (\bibinfo{year}{2017}).
\newblock


\bibitem[Wen et~al\mbox{.}(2021)]%
        {Wen21}
\bibfield{author}{\bibinfo{person}{Yandong Wen}, \bibinfo{person}{Weiyang Liu},
  \bibinfo{person}{Bhiksha Raj}, {and} \bibinfo{person}{Rita Singh}.}
  \bibinfo{year}{2021}\natexlab{}.
\newblock \showarticletitle{Self-supervised 3d face reconstruction via
  conditional estimation}. In \bibinfo{booktitle}{\emph{Proceedings of the
  IEEE/CVF International Conference on Computer Vision}}.
  \bibinfo{pages}{13289--13298}.
\newblock


\bibitem[Zhang et~al\mbox{.}(2023)]%
        {zhang2023dreamface}
\bibfield{author}{\bibinfo{person}{Longwen Zhang}, \bibinfo{person}{Qiwei Qiu},
  \bibinfo{person}{Hongyang Lin}, \bibinfo{person}{Qixuan Zhang},
  \bibinfo{person}{Cheng Shi}, \bibinfo{person}{Wei Yang}, \bibinfo{person}{Ye
  Shi}, \bibinfo{person}{Sibei Yang}, \bibinfo{person}{Lan Xu}, {and}
  \bibinfo{person}{Jingyi Yu}.} \bibinfo{year}{2023}\natexlab{}.
\newblock \showarticletitle{Dreamface: Progressive generation of animatable 3d
  faces under text guidance}.
\newblock \bibinfo{journal}{\emph{arXiv preprint arXiv:2304.03117}}
  (\bibinfo{year}{2023}).
\newblock


\bibitem[Zhang et~al\mbox{.}(2018)]%
        {zhang2018perceptual}
\bibfield{author}{\bibinfo{person}{Richard Zhang}, \bibinfo{person}{Phillip
  Isola}, \bibinfo{person}{Alexei~A Efros}, \bibinfo{person}{Eli Shechtman},
  {and} \bibinfo{person}{Oliver Wang}.} \bibinfo{year}{2018}\natexlab{}.
\newblock \showarticletitle{The unreasonable effectiveness of deep features as
  a perceptual metric}. In \bibinfo{booktitle}{\emph{Proceedings of the IEEE
  conference on computer vision and pattern recognition}}.
  \bibinfo{pages}{586--595}.
\newblock


\bibitem[Zhu et~al\mbox{.}(2021)]%
        {zhu2021webface260m}
\bibfield{author}{\bibinfo{person}{Zheng Zhu}, \bibinfo{person}{Guan Huang},
  \bibinfo{person}{Jiankang Deng}, \bibinfo{person}{Yun Ye},
  \bibinfo{person}{Junjie Huang}, \bibinfo{person}{Xinze Chen},
  \bibinfo{person}{Jiagang Zhu}, \bibinfo{person}{Tian Yang},
  \bibinfo{person}{Jiwen Lu}, \bibinfo{person}{Dalong Du}, {et~al\mbox{.}}}
  \bibinfo{year}{2021}\natexlab{}.
\newblock \showarticletitle{Webface260m: A benchmark unveiling the power of
  million-scale deep face recognition}. In
  \bibinfo{booktitle}{\emph{Proceedings of the IEEE/CVF Conference on Computer
  Vision and Pattern Recognition}}. \bibinfo{pages}{10492--10502}.
\newblock


\end{thebibliography}

\begin{figure*}[t!]
\centering
\includegraphics[width=0.98\linewidth]{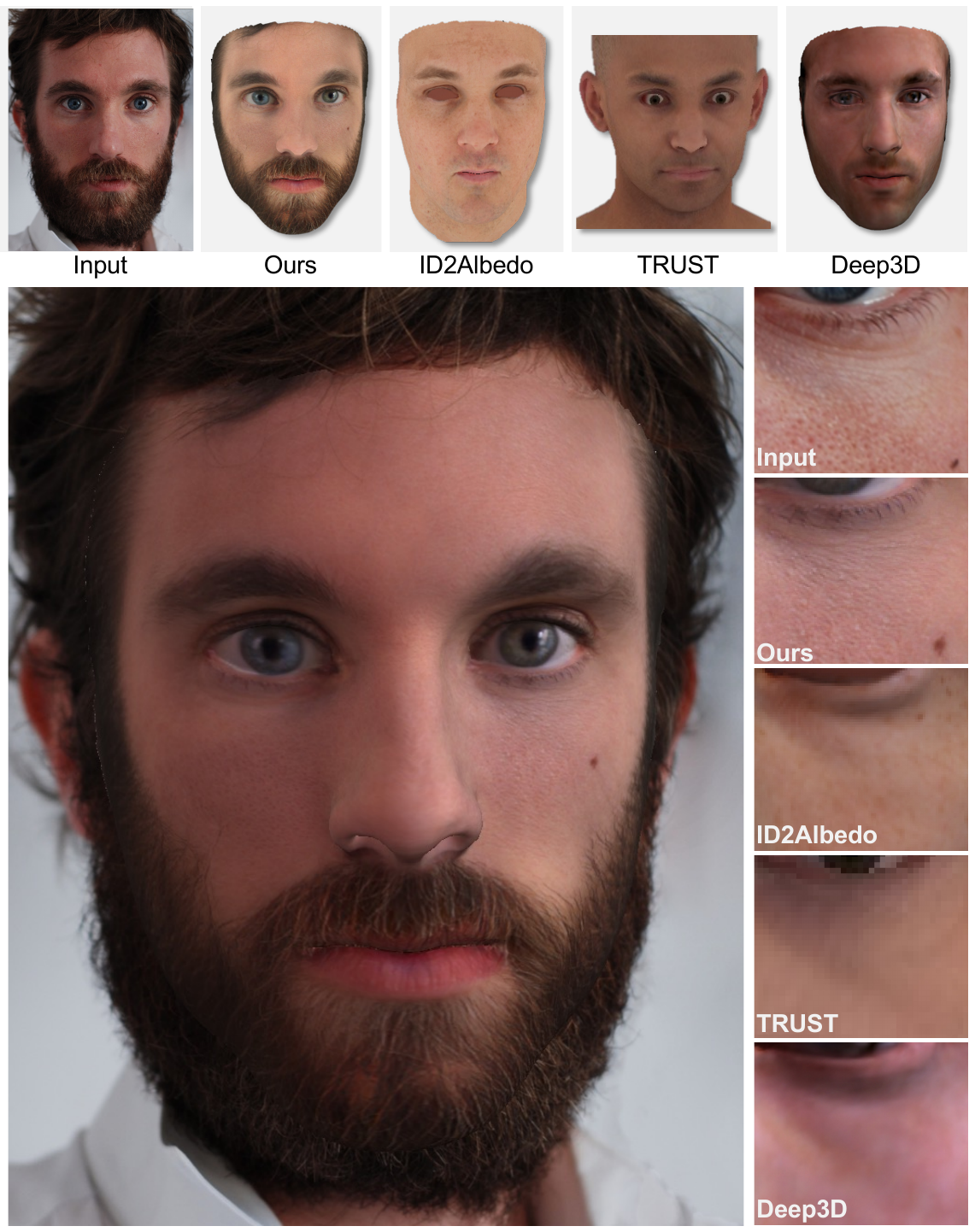}
\caption{Comparisons on albedo details with ID2Albedo~\cite{ren2023improving}, TRUST~\cite{feng2022towards} and Deep3D~\cite{deng2019accurate} albedo and rendered images. We achieve the most realistic rendered results and our generated high-fidelity albedo preserves the skin texture, pores, and moles on the face.}
\label{fig:figure-only}
\end{figure*}

\begin{figure*}[t!]
\centering
\includegraphics[width=\textwidth]{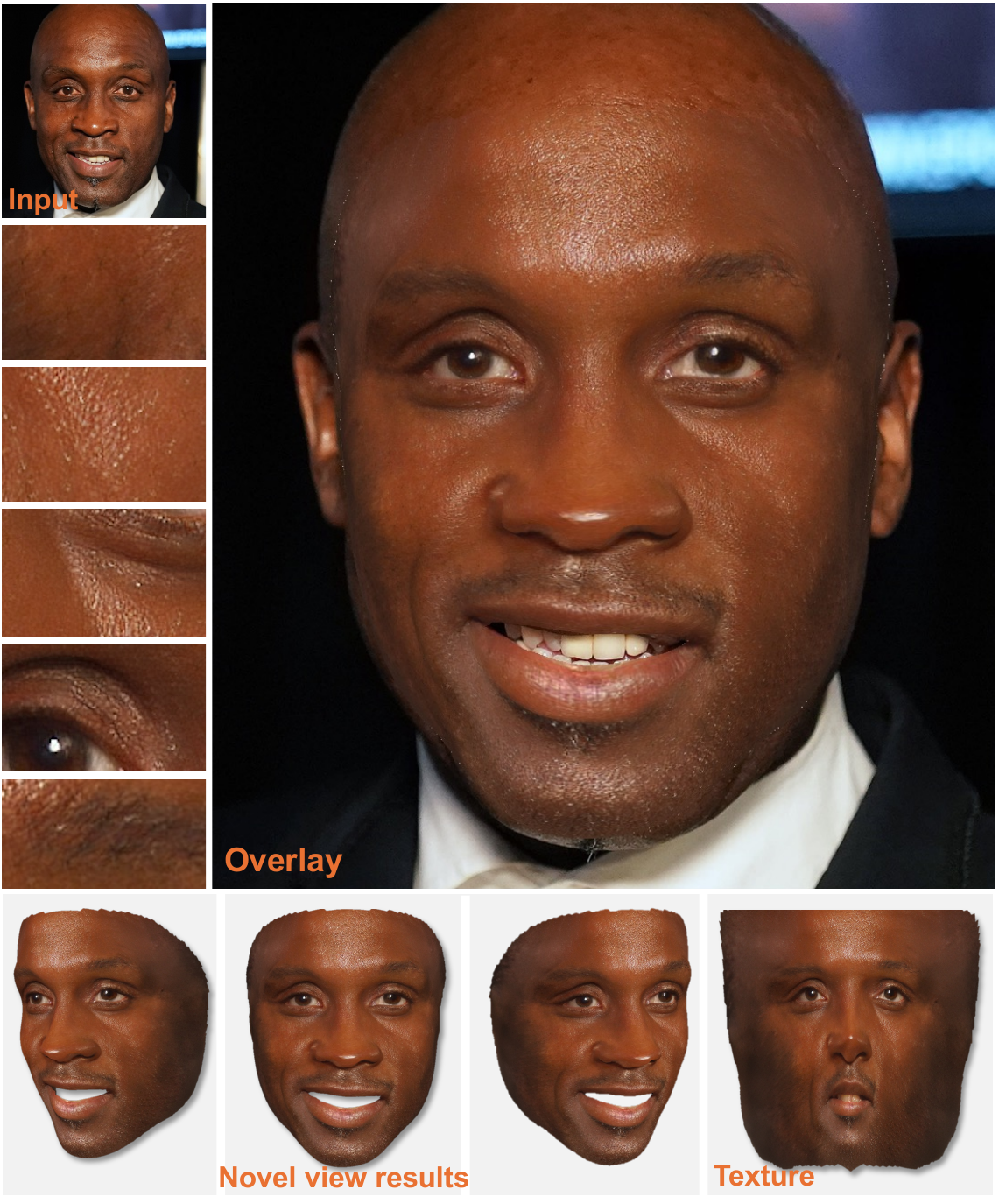}
\caption{Visualization of texture reconstruction. Given a single image as input, our method reconstructs extremely high-quality texture details.}
\label{fig:figure-only}
\end{figure*}

\clearpage
\appendix

\section*{Supplemental Materials}

In this supplementary material, 
we first introduce datasets we utilized in the main manuscript (Sec.~\ref{dataset}).
Then, we introduce more ablation studies to better evaluate the modules in HiFiAlbedo (Sec.~\ref{ablation}).
Finally, we discuss the limitations of HiFiAlbedo (Sec.~\ref{limitation}) and then provide more qualitative results to demonstrate the robustness of the proposed method (Sec.~\ref{qualitative}).

\section{Datasets}\label{dataset}
\begin{figure}[b]
\centering
\includegraphics[width=0.98\linewidth]{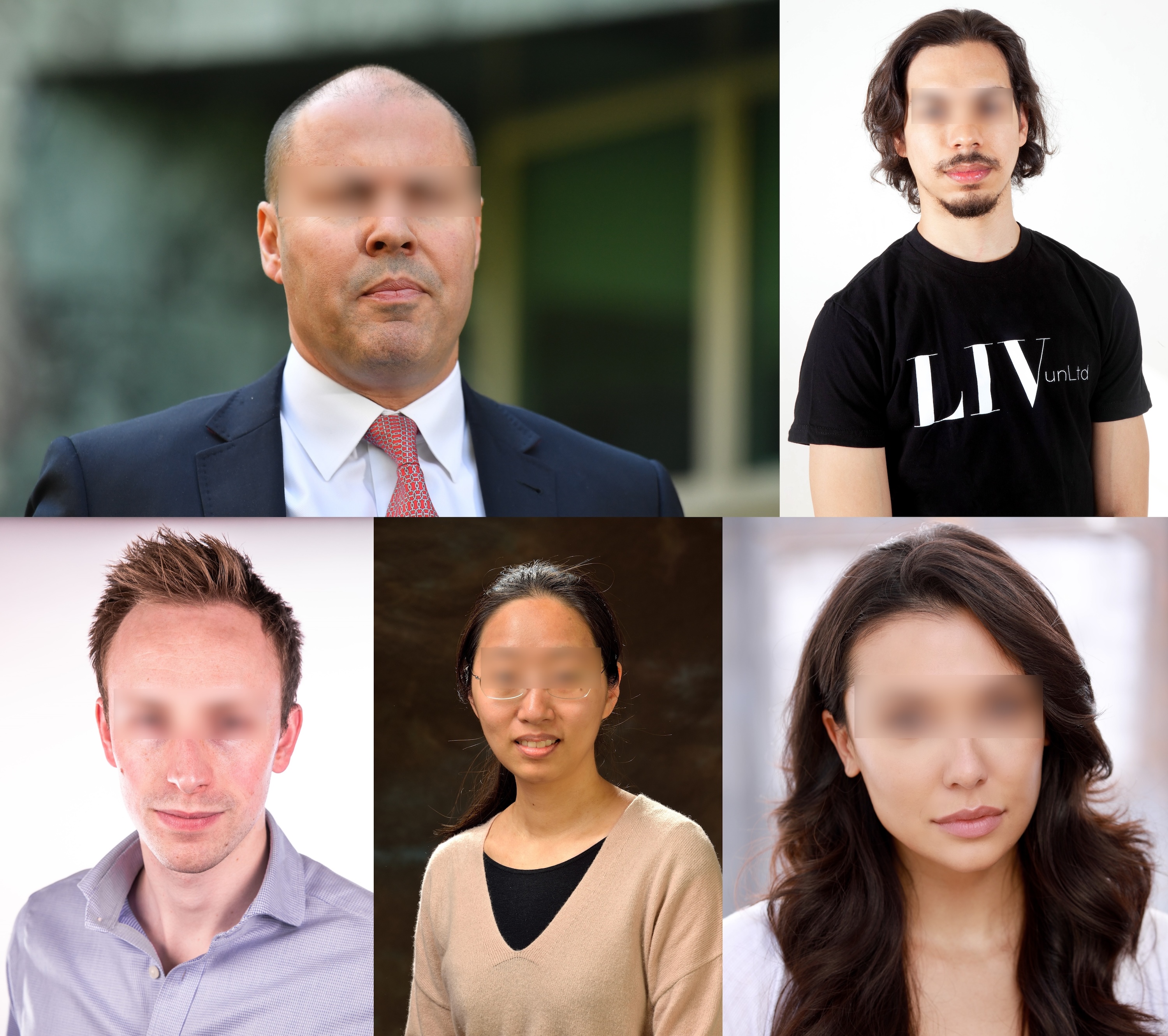}
\caption{Examples of our large-scale face dataset, sourced from WebFace260M~\cite{zhu2021webface260m}. Our images contain a lot of skin detail on the face.}
\label{suppfig:webface}
\end{figure}

To create a high-quality facial texture codebook, we gathered a large-scale, high-quality face dataset comprising 500,000 images. Example images are provided in Fig.~\ref{suppfig:webface} and the statistical results are shown in Fig.~\ref{dataset}. Our high-fidelity dataset is derived from the large-scale open-source face dataset, WebFace260M~\cite{zhu2021webface260m}. We further utilized the open-source face processing toolkit InsightFace~\cite{insightface2019} to filter out high-quality face images which bounding box exceed a resolution of 1024. During the model training phase, we initially crop the faces and then resize them to $1024\times1024$ for training. We will make this data open-source to enable more researchers and developers to utilize it and advance related fields.

\begin{figure*}[t!]
\centering
\includegraphics[width=0.98\linewidth]{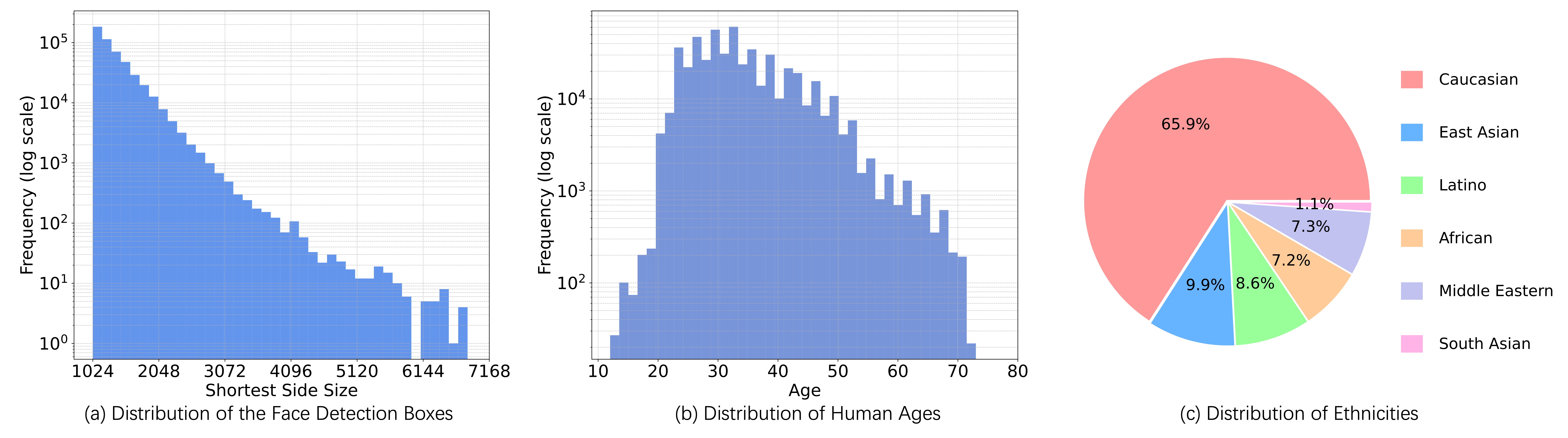}
\caption{Overview of our dataset. We show (a) face size distribution (b) age distribution and (c) ethnicity distribution in the dataset, respectively.}
\label{suppfig:dataset}
\end{figure*}

\section{More Ablation Studies}\label{ablation}
\noindent{\bf High-Fidelity Codebook.}
Tab.~\ref{tab:abl:reconstruction} shows that our high-fidelity codebook demonstrates strong reconstruction capabilities in terms of face, texture, and albedo. It achieves high PSNR, SSIM, and LPIPS scores, significantly outperforming the original VQGAN~\cite{esser2021taming}. Moreover, as depicted in Fig.~\ref{fig:3_1}, the reconstructed outputs exhibit considerable detail restoration in face detail, texture, and albedo. This suggests that our codebook can effectively reconstruct high-fidelity attributes in these different aspects.
\begin{table}[t]
\centering
\resizebox{\linewidth}{!}{%
\begin{tabular}{l|cccc}
    \toprule
    Method &  Dataset (Domain) & PSNR$\uparrow$ & SSIM$\uparrow$   &LPIPS$\downarrow$  \\ 
    \midrule
    VQGAN         & CelebAMask-HQ (Face)    & 27.05  & 0.8367 & 0.2957 \\ 
    Ours               & CelebAMask-HQ (Face)    & 28.60  & 0.8460 & 0.2451 \\ 
    \hline
    VQGAN         & OSTeC (Texture) & 28.39 & 0.8945 & 0.3590 \\ 
    Ours               & OSTeC (Texture) & 31.88 & 0.9007 & 0.2879 \\ 
    \hline
    VQGAN         & 3DScan Store (Albedo) & 29.49  & 0.8122 & 0.3411 \\ 
    Ours               & 3DScan Store (Albedo) & 31.00  & 0.7634 & 0.2454 \\ 
    \bottomrule
\end{tabular}}
\caption{Comparison of reconstruction capabilities with VQGAN~\cite{esser2021taming} across CelebAMask-HQ (Face), OSTeC (Texture), and 3DScan Store (Albedo).}
\label{tab:abl:reconstruction}
\end{table}

\begin{figure*}[t!]
\centering
\includegraphics[width=0.98\linewidth]{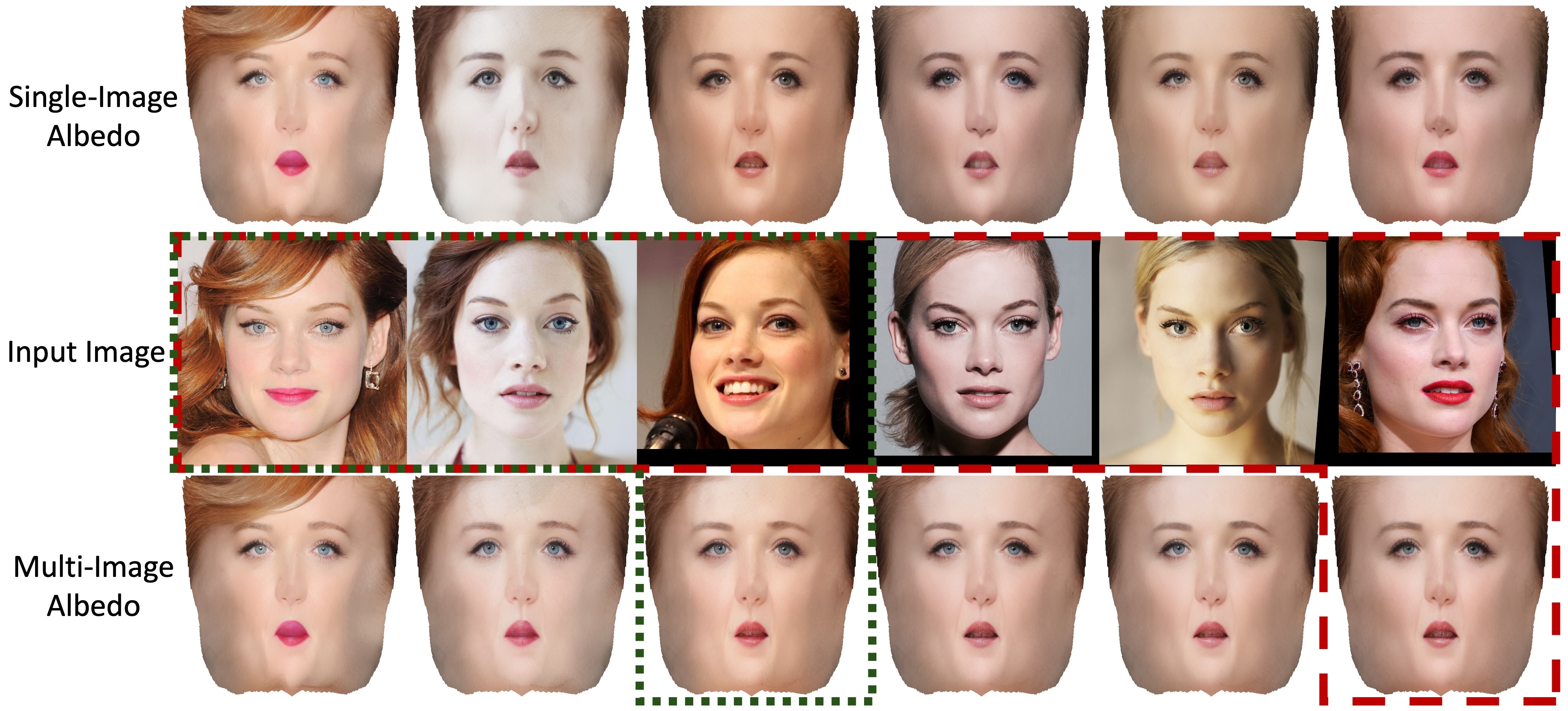}
\caption{Ablation study of multi-image inference. The first row displays single input image albedo maps, while the last row shows increasing multi input images generated albedo.}
\label{fig:multi-infer}
\end{figure*}

\begin{figure*}[t!]
\centering
\includegraphics[width=0.98\linewidth]{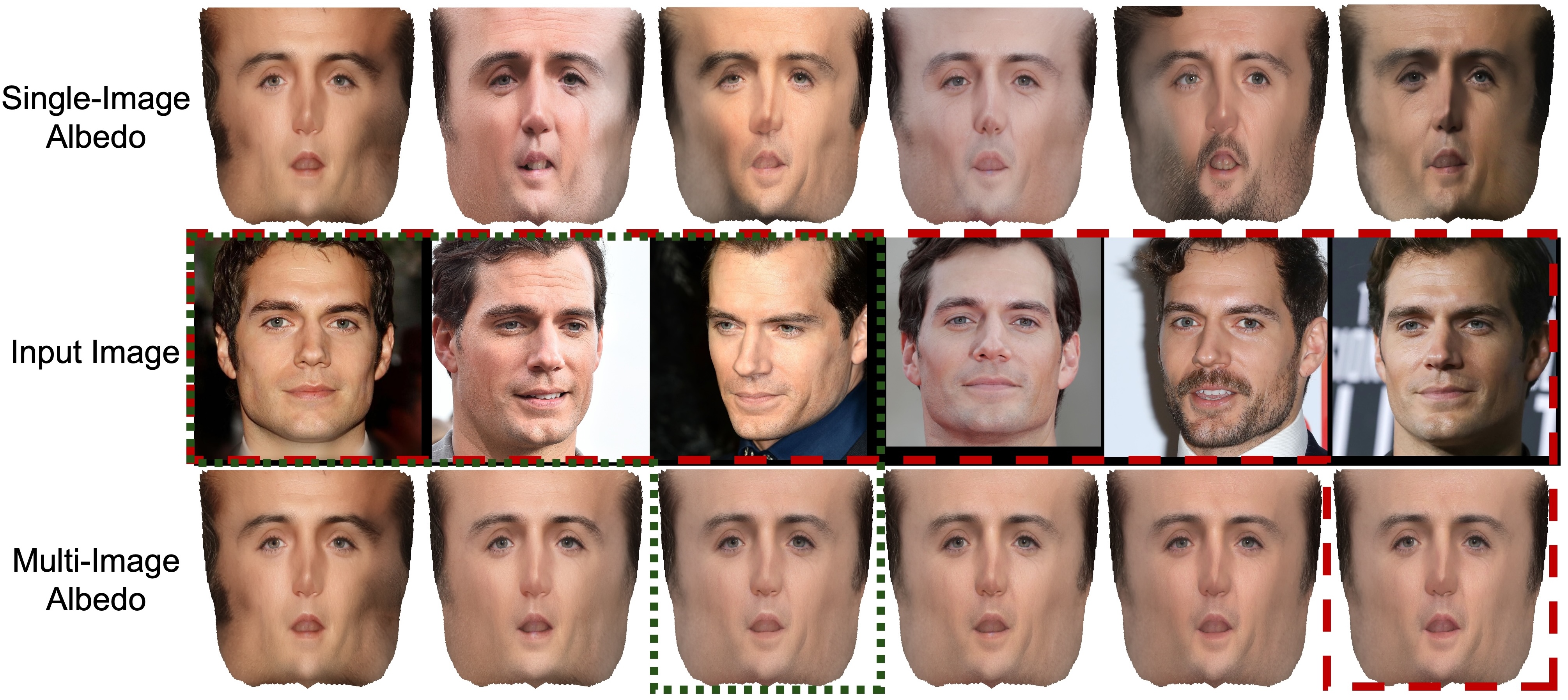}
\caption{Ablation study of multi-image inference. The first row displays single input image albedo maps, while the last row shows increasing multi input images generated albedo.}
\label{fig:multi-infer2}
\end{figure*}

\begin{figure*}[t!]
\centering
\includegraphics[width=0.98\linewidth]{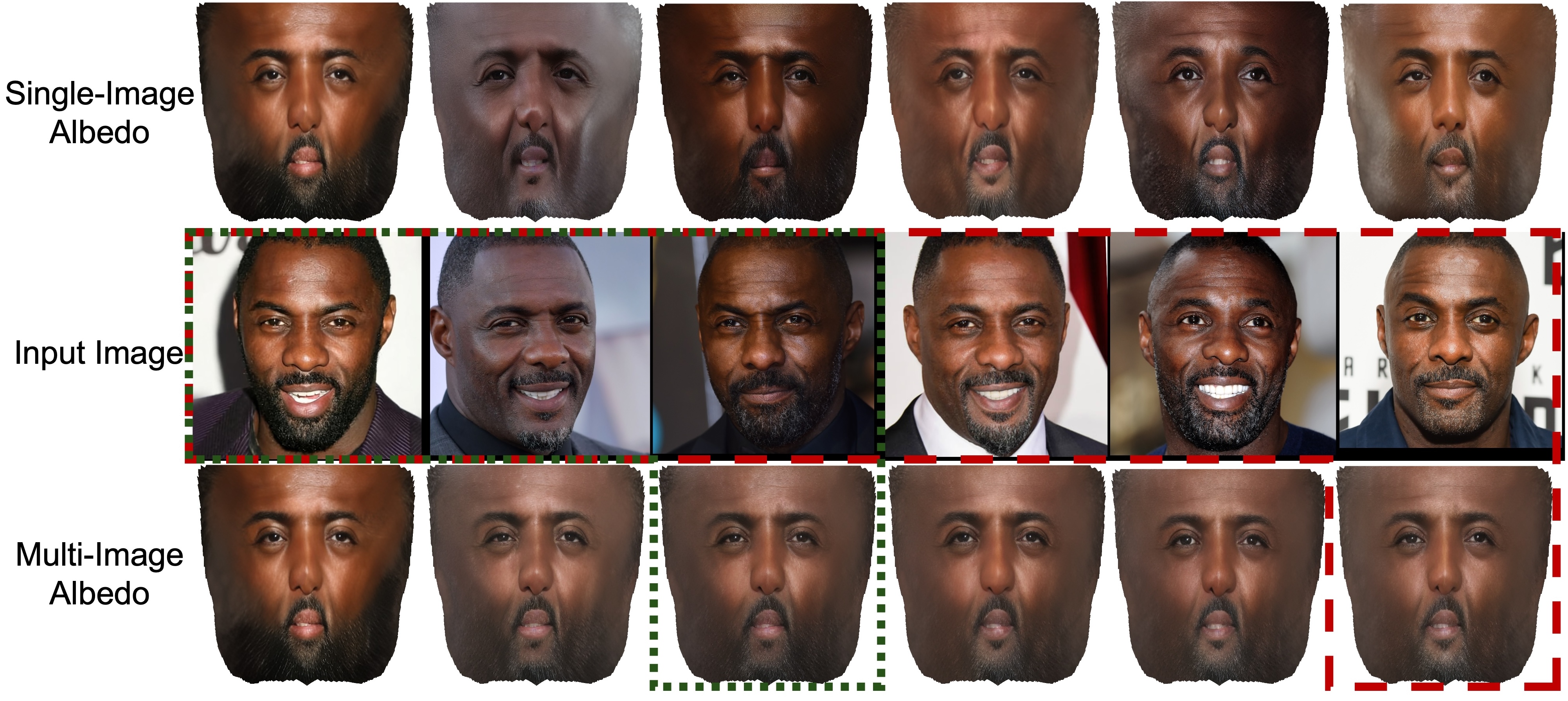}
\caption{Ablation study of multi-image inference. The first row displays single input image albedo maps, while the last row shows increasing multi input images generated albedo.}
\label{fig:multi-infer3}
\end{figure*}

\noindent{\bf Multi-image Inference.}
We further provide more results to demonstrate the advantages of multi-image inference. As shown in the first row of Fig.~\ref{fig:multi-infer},~\ref{fig:multi-infer2}, and~\ref{fig:multi-infer3}, the model's inference for single images is influenced by albedo/illumination blur. However, the albedo inference quality for multiple images improves with increasing input images, as observed in the last row. The green boxes depict albedo maps derived from the first three inputs, while the red boxes represent maps obtained from all six inputs.

\section{Limitations}\label{limitation}
Given that our method is self-supervised and utilizes only in-the-wild images, this data-driven approach presents several challenges.
Firstly, within the UV texture reconstruction phase, our method is influenced by the input image resolution and occlusions. This can be mitigated through pre-processing of the input image.
Secondly, for the albedo reconstruction part, our method necessitates diversity in the input images. In other words, if only a few images from the same scene are provided, the inference results will degrade to the equivalent of a single image input.
Lastly, due to the data-driven nature of our method, we are unable to entirely separate the lighting from the images, leaving some remnants of artifacts. We acknowledge the difficulty in flawlessly recovering albedo without using Ground Truth data and believe that improved light estimation results and light modeling methods offer a viable pathway forward.

\section{More Qualitative Results}\label{qualitative}
\begin{figure*}[t!]
\centering
\includegraphics[width=0.98\linewidth]{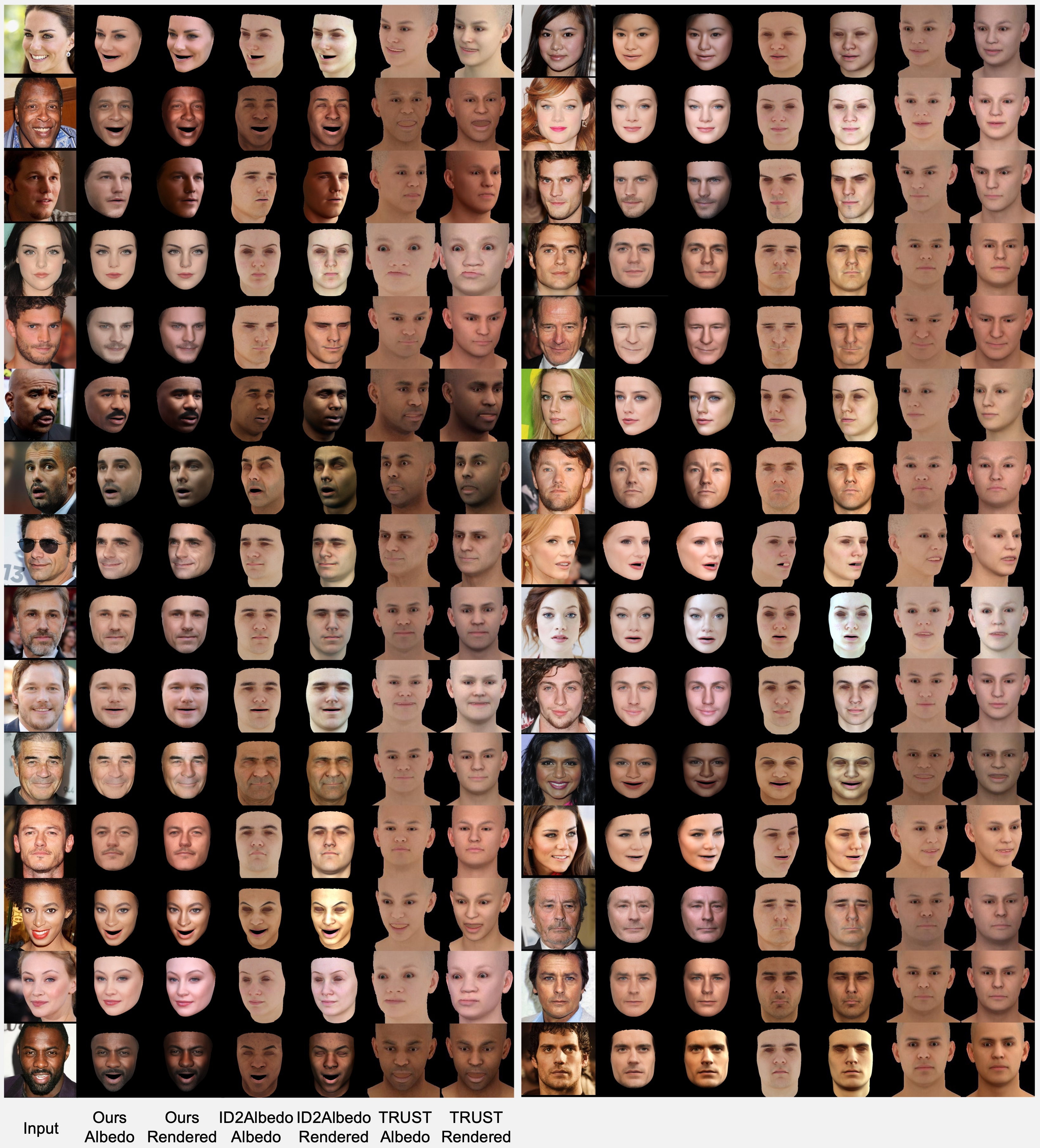}
\caption{Comparisons on albedo details with ID2Albedo~\cite{ren2023improving} and TRUST~\cite{feng2022towards} albedo and rendered images. From left to right are the input image, multi-image inference albedo, rendered images. We use the existing Webface260M~\cite{zhu2021webface260m} to find additional 3 photos of the same person for inference.}
\label{suppfig:albedo1}
\end{figure*}

\begin{figure*}[t!]
\centering
\includegraphics[width=0.98\linewidth]{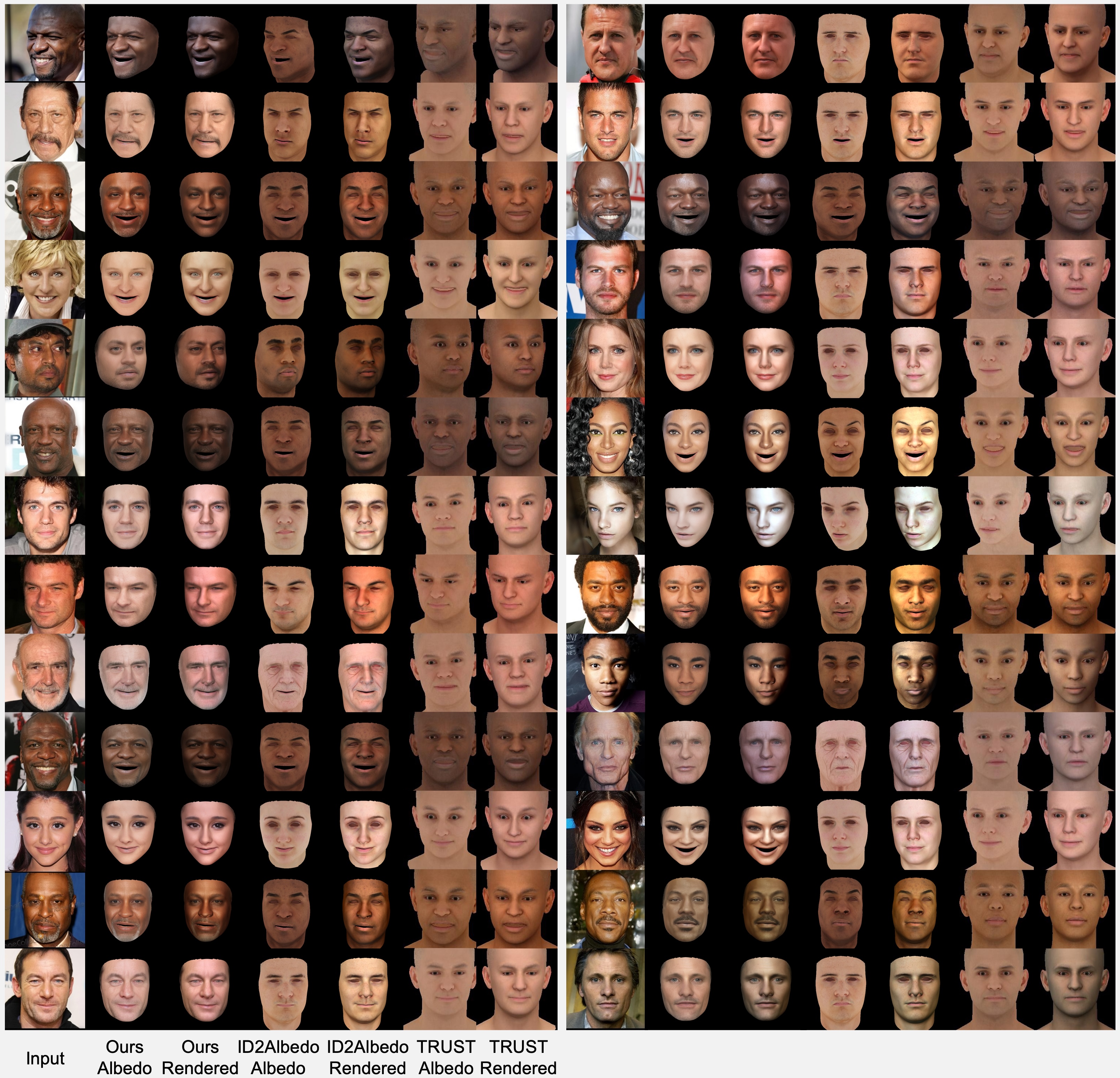}
\caption{Comparisons on albedo details with ID2Albedo~\cite{ren2023improving} and TRUST~\cite{feng2022towards} albedo and rendered images. From left to right are the input image, multi-image inference albedo, rendered images. We use the existing Webface260M~\cite{zhu2021webface260m} to find additional 3 photos of the same person for inference.}
\label{suppfig:albedo2}
\end{figure*}

\begin{figure*}[t!]
\centering
\includegraphics[width=0.95\linewidth]{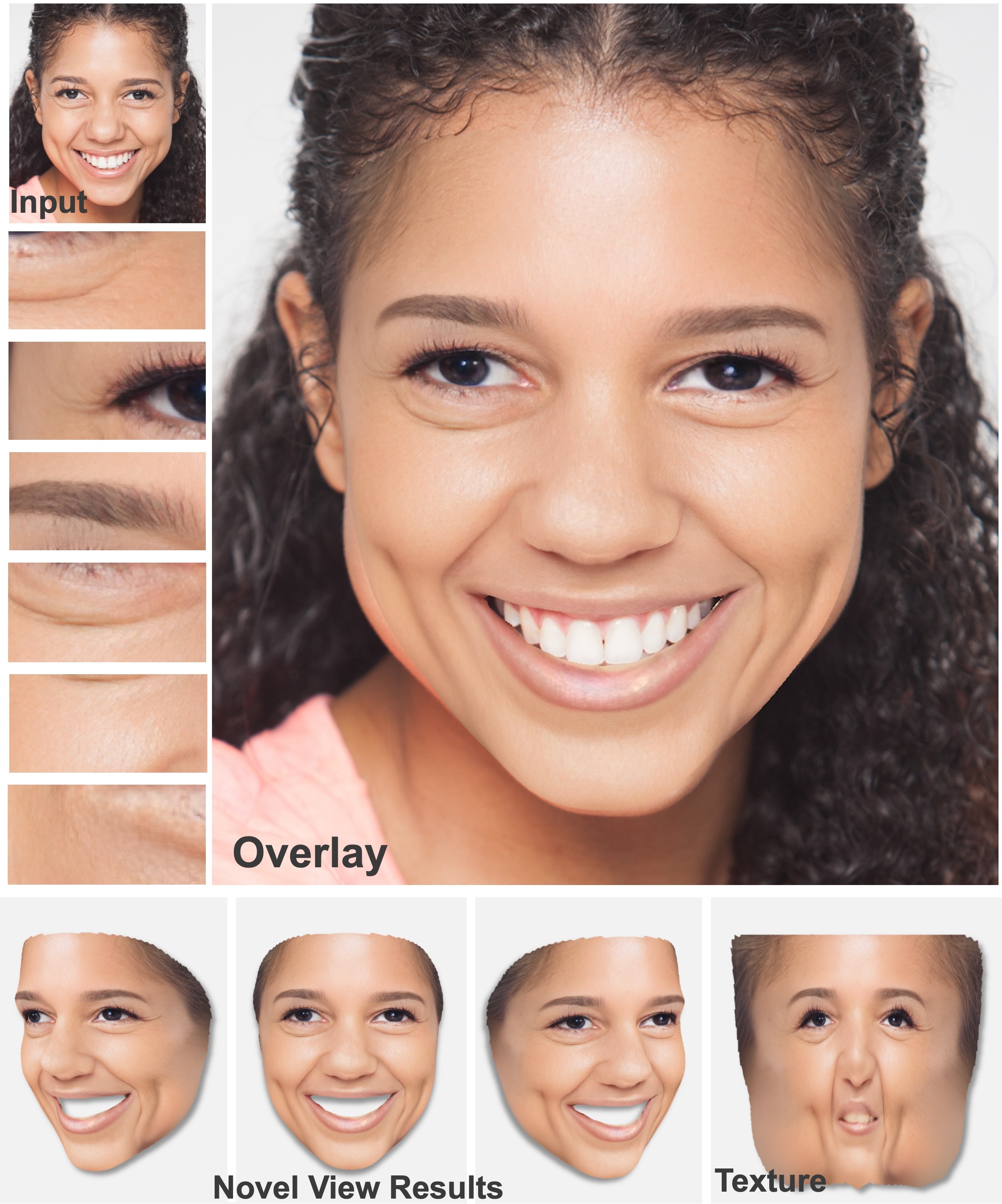}
\caption{Visualization of texture reconstruction. Given a single image as input, our method reconstructs extremely high-quality texture details.}
\label{suppfig:supp_tex1}
\end{figure*}

\begin{figure*}[t!]
\centering
\includegraphics[width=0.95\linewidth]{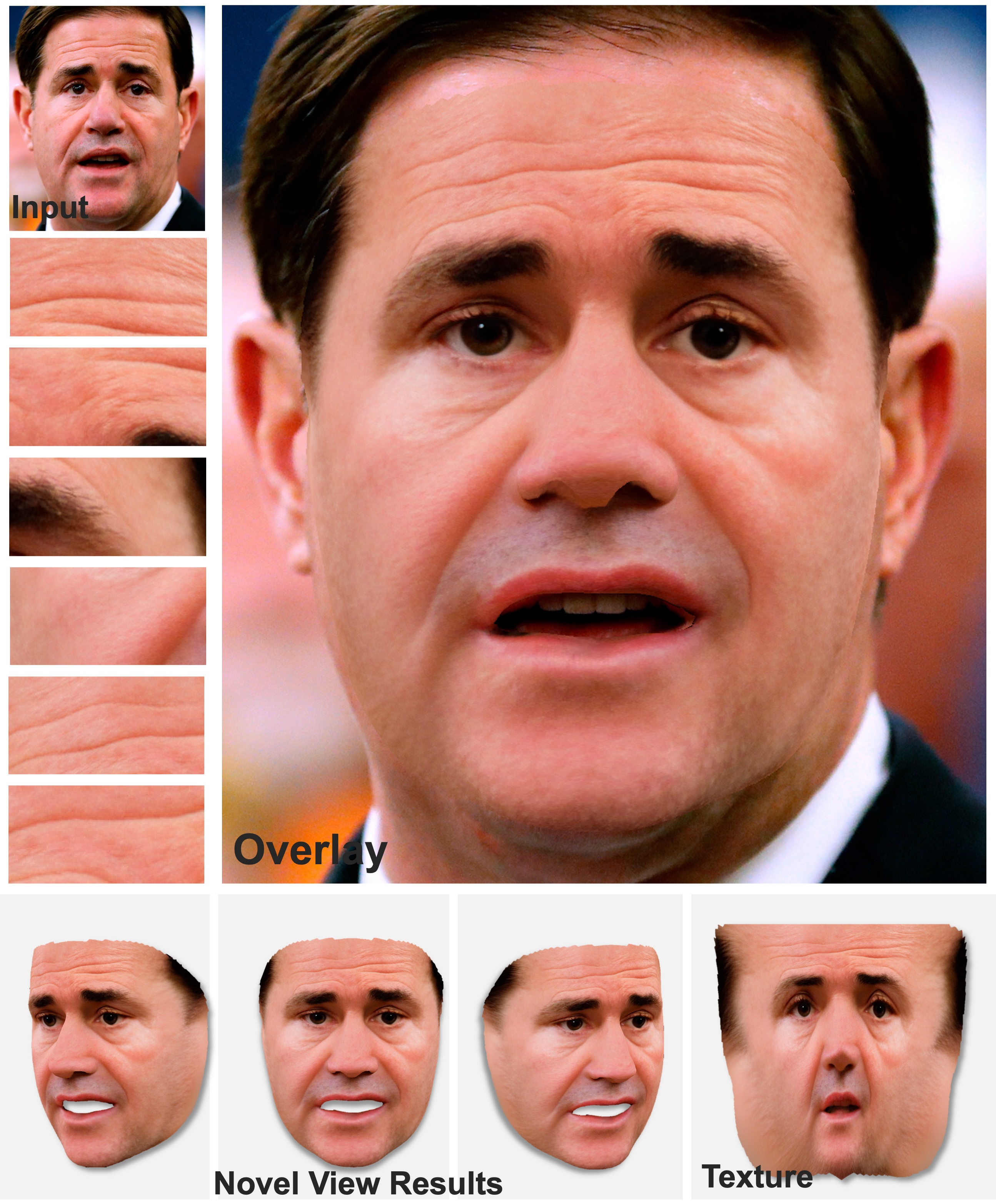}
\caption{Visualization of texture reconstruction. Given a single image as input, our method reconstructs extremely high-quality texture details.}
\label{suppfig:supp_tex2}
\end{figure*}

\begin{figure*}[t!]
\centering
\includegraphics[width=0.95\linewidth]{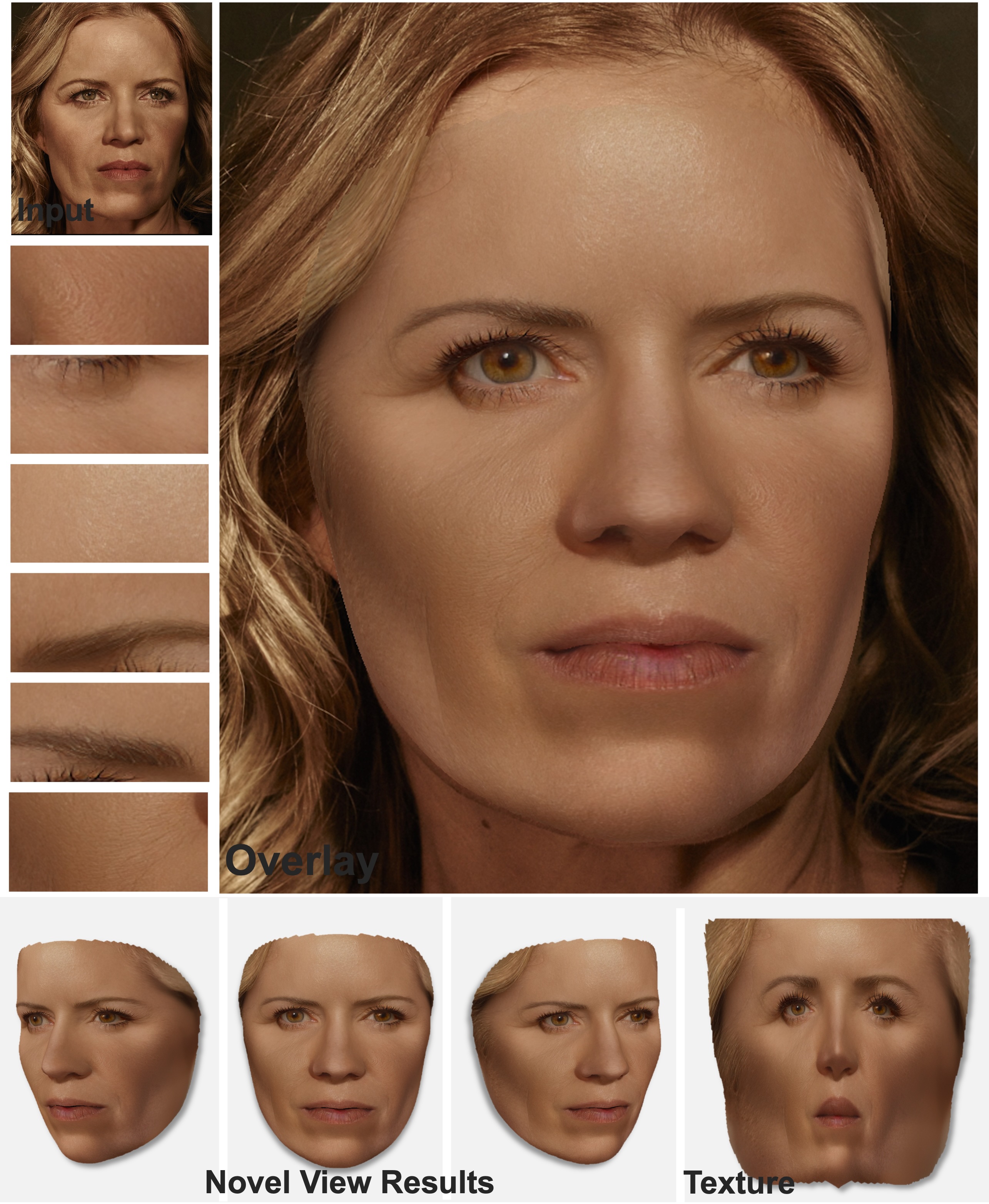}
\caption{Visualization of texture reconstruction. Given a single image as input, our method reconstructs extremely high-quality texture details.}
\label{suppfig:supp_tex3}
\end{figure*}

\begin{figure*}[t!]
\centering
\includegraphics[width=0.95\linewidth]{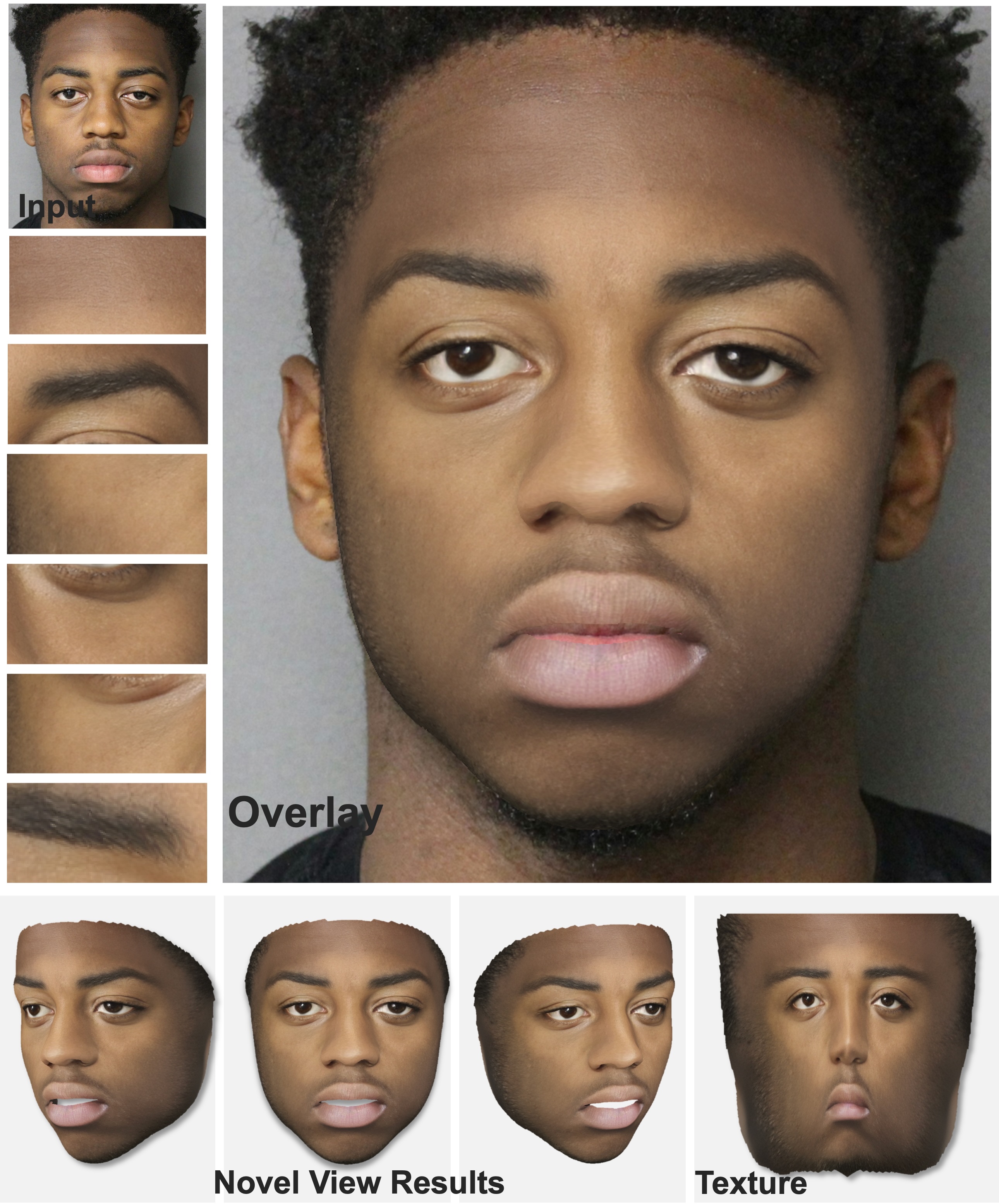}
\caption{Visualization of texture reconstruction. Given a single image as input, our method reconstructs extremely high-quality texture details.}
\label{suppfig:supp_tex4}
\end{figure*}

We provide more albedo reconstruction results for comparison with TRUST~\cite{feng2022towards} and ID2Albedo~\cite{ren2023improving}. The results in Fig.~\ref{suppfig:albedo1} and~\ref{suppfig:albedo2} show that our albedo achieves the highest fidelity while maintaining comparable fairness.
More details of the UV texture reconstruction are also shown in Fig.~\ref{suppfig:supp_tex1},~\ref{suppfig:supp_tex2},~\ref{suppfig:supp_tex3} and ~\ref{suppfig:supp_tex4}.

\end{document}